\documentclass[review]{elsarticle}

\usepackage{lineno,hyperref}
\modulolinenumbers[5]

\usepackage{geometry}
\journal{Journal of Information Sciences}

\usepackage{times}
\usepackage{amsmath}
\usepackage{soul}
\usepackage{url}
\usepackage[utf8]{inputenc}
\usepackage{bbding}
\usepackage{graphicx}
\usepackage{amsmath}
\usepackage{booktabs}
\usepackage{framed}
\usepackage{algorithm}
\usepackage{algorithmic}
\usepackage{lineno}
\usepackage{pifont}
\usepackage{graphicx}
\usepackage{subfigure}
\usepackage{multirow}
\usepackage{amsmath}
\usepackage{amsfonts}
\usepackage{color}
\usepackage{float}

\graphicspath{ {images/} }

\begin{document}

\begin{frontmatter}

\title{Salient Feature Extractor for Adversarial Defense on \\ Deep Neural Networks}

\author[mymainaddress,mysecondaryaddress]{Jinyin Chen\corref{mycorrespondingauthor}}
\cortext[mycorrespondingauthor]{Corresponding author}
\ead{chenjinyin@zjut.edu.cn}

\author[mysecondaryaddress]{Ruoxi Chen}
\ead{2112003149@zjut.edu.cn}

\author[mysecondaryaddress]{Haibin Zheng}
\ead{haibinzheng320@gmail.com}

\author[mythirdaddress]{Zhaoyan Ming}
\ead{mingzhaoyan@gmail.com}

\author[myfourthaddress]{Wenrong Jiang}
\ead{jiangwenrong@zjjcxy.com}

\author[myfifthaddress]{Chen Cui}
\ead{cuichen@zjjcxy.com}

\address[mymainaddress]{Institute of Cyberspace Security, Zhejiang University of Technology, Hangzhou, China}
\address[mysecondaryaddress]{College of Information Engineering, Zhejiang University of Technology, Hangzhou, China}

\address[mythirdaddress]{Institute of Computing Innovation, Zhejiang Univeristy, Hangzhou, China}
\address[myfourthaddress]{College of Computer Science, Hangzhou Dianzi University,Hangzhou, China}
\address[myfifthaddress]{the Big Data and Cyber Security Research Institute, Zhejiang
Police College, Hangzhou, China}

\begin{abstract}
Recent years have witnessed unprecedented success achieved by deep learning models in the field of computer vision. However, their vulnerability towards carefully crafted adversarial examples has also attracted the increasing attention of researchers. Motivated by the observation that adversarial examples are due to the non-robust feature learned from the original dataset by models, we propose the concepts of salient feature(SF) and trivial feature(TF). The former represents the class-related feature, while the latter is usually adopted to mislead the model. We extract these two features with coupled generative adversarial network model and put forward a novel detection and defense method named salient feature extractor (SFE) to defend against adversarial attacks. Concretely, detection is realized by separating and comparing the difference between SF and TF of the input. At the same time, correct labels are obtained by re-identifying SF to reach the purpose of defense. Extensive experiments are carried out on MNIST, CIFAR-10, and ImageNet datasets where SFE shows state-of-the-art results in effectiveness and efficiency compared with baselines. Furthermore, we provide an interpretable understanding of the defense and detection process. The code of SFE could be downloaded from \url{https://github.com/haibinzheng/SFE} .

\end{abstract}


\end{frontmatter}

% \linenumbers

\section{Introduction}
Deep learning enjoys great popularity in both academic and industrial application for its superior performance, ranging from the field of image classification to object detection, natural language processing and bioinformatics analysis.
However, deep neural networks (DNNs) are vulnerable to adversarial perturbations, imperceptible to humans, easily lead to misclassification, as proved by Szegedy~\cite{szegedy2013intriguing}. In the process of independent decision-making, the vulnerability of deep models to adversarial examples has posed non-negligible threat to data and information security. This problem, furthermore, impedes the application in mission-critical areas, such as face recognition and auto pilot. Therefore, it is crucial to study defense against adversarial attacks and further improve the robustness of deep models.

In the field of image classification, numerous adversarial attacks have been proposed to discover vulnerabilities of DNNs. Based on the knowledge degree of the target model, adversarial attacks are categorized into white-box attacks and black-box attacks. One of the typical white-box attacks is gradient-based attack, e.g., fast gradient sign method (FGSM)~\cite{goodfellow2014explaining}, basic iterative method (BIM)~\cite{Kurakin2017Adversarial}, momentum-based iterative FGSM (MI-FGSM)~\cite{Dong2018Boosting} and decision boundary-based attack DeepFool~\cite{moosavi2016deepfool}. Those attacks require less time to compute perturbations but a thorough knowledge of the targeted model should be given in advance. Black-box attacks are mainly decision-based and score-based, such as the zeroth-order optimization attack (ZOO)~\cite{chen2017zoo}, point wise attack (PWA)~\cite{schott2018towards} and local search attack (LSA)~\cite{narodytska2016simple}. Only knowing the output of the targeted model, black-box attacks can still be carried out, usually with a larger size of perturbation. 

Meanwhile, the profound implications of DNN’s vulnerability have motivated a wide range of investigations into defense for DNNs. Depending on different defense purposes, they can be categorized as re-identification defense and adversarial detector. Re-identification defenses, known as complete defenses, are now developed along three main directions. Training/input modification includes adversarial training~\cite{goodfellow2014explaining} and input pre-processing; model modification includes defense distillation~\cite{papernot2016distillation} and reverse cross-entropy loss~\cite{pang2018towards}; defense of network add-on contains methods based on generative adversarial network(GAN)~\cite{goodfellow2014generative} and autoencoder~\cite{hlihor2020evaluating}. Re-identification defense can provide correct class labels of adversarial examples while adversarial detector only determines whether the input is benign or adversarial. The ACT-detector~\cite{chen2021act}, perturbation detection~\cite{metzen2017detecting}, GAT~\cite{Yin2020GAT} are three of those available defenses. 

\begin{figure}[t]
  \centering
  \subfigure[benign example]{
  \includegraphics[width=0.19\linewidth]{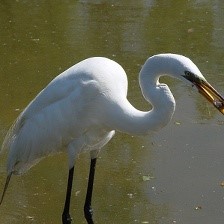} }
  \subfigure[heatmap of benign]{
  \includegraphics[width=0.19\linewidth]{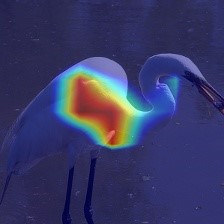} }
  \subfigure[SF of benign]{
  \includegraphics[width=0.19\linewidth]{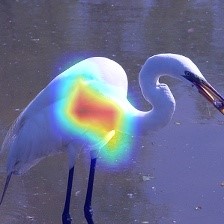} }
  \subfigure[TF of benign]{
  \includegraphics[width=0.19\linewidth]{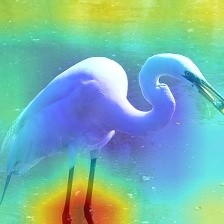} }
  \\
  \subfigure[adversarial example]{
  \includegraphics[width=0.19\linewidth]{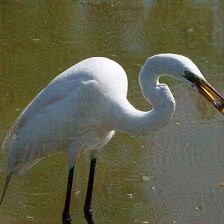} }
  \subfigure[heatmap of adversarial]{
  \includegraphics[width=0.19\linewidth]{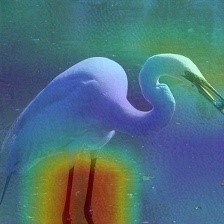} }
  \subfigure[SF of adversarial]{
  \includegraphics[width=0.19\linewidth]{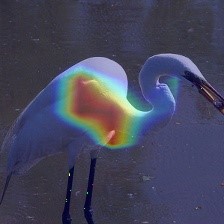} }
  \subfigure[TF of adversarial]{
  \includegraphics[width=0.19\linewidth]{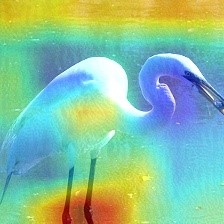} }
  \caption{Visualization of SF and TF for VGG19 model on ImageNet via Grad-CAM. Red area in SF of adversarial looks similar to that in benign heatmaps, leading to successful defense. Four images from left to right are the benign example (classified as "American egret") and its heatmap, visualization of SF and TF reconstructed by SFE. The second row represents heatmaps of adversarial examples (misclassified as " cheetah") generated by MI-FGSM and its corresponding SF and TF results after reconstruction. Heatmaps in the second column uses the logits of benign input for calculation while the last two columns use that of the extracted SF and TF.  }
  \label{fig:ex_heatmaps}
\end{figure}

Another novel angle to study the adversarial example is from MIT~\cite{ilyas2019adversarial}, they first claimed that the adversarial example is not a bug, concluding that the existence of adversarial examples arise from the non-robust features learned from the original dataset by the model. They gave interpretation of adversarial example from the machine's angle instead of man. From that point of view, the adversarial example takes advantage of non-robust feature to fool DNN while the robust feature is still working for man since the adversarial perturbation is imperceptible. Inspired by this, we propose the concepts of salient feature(SF) and trivial feature(TF), then compare visualization of these features for benign and adversarial examples via Grad-CAM~\cite{selvaraju2017grad}, heatmaps on ImageNet VGG19~\cite{simonyan2014very} model shown in Figure\ref{fig:ex_heatmaps}. From red to blue, the weight that the model allocates decreases. By comparing heatmaps, we find that the red area of SF in the benign example is similar to the heatmap of original map, but different from that of the adversarial. Correct labels, consistent with human prediction, can be gained when the model pays attention to SF. On the contrary, misclassification occurs when TF is focused.

Since the SF captures the salient features of the adversarial examples, we proposed a novel defense method, salient feature extractor, SFE for short. It extracts and reconstructs SF and TF to defend against adversarial attacks. It has been proved that generative adversarial network (GAN)~\cite{goodfellow2014generative} is a well-designed optimizer on basis of game player. We adopt a coupled GAN model to extract SF and TF respectively. Adversarial examples are distinguished by calculating difference between SF and TF. Consequently, correct classification labels are obtained by recognition of reconstructed and strengthened SF in adversarial examples.

Our main contributions are summarized as follows:
\begin{enumerate}
  \item We propose concepts of salient features(SF) and trivial features(TF). Observation is obtained that the correct classification of examples attributes to the model's concentration on SF. Moreover, consistent with human semantics, SF and TF could be adopted for detection and defense strategy.  
  \item We design a feature extraction method of SF and TF, namely SFE, which can effectively achieve the detection of benign and adversarial examples by feature separation and calculation of distribution difference. By reconstructing SF and TF of adversarial examples on the basis of coupled GAN framework, both detection and re-identification defense of adversarial examples can be completed.  
  \item Extensive experiments are implemented on various models and datasets to demonstrate the detection and defense effect of SFE, including within and among-class distance, transferability, parameter sensitivity and time complexity. Besides, interpretable defense via visualization is provided for better understanding.
\end{enumerate}

The rest of the paper is organized as follows.
The related works are discussed in Section~\ref{RWs}, the SFE method and critical techniques are introduced in Section~\ref{method}. Experiments and analysis are detailed in Section~\ref{Exp} respectively. At last, we conclude limitations and future works.

\section{Related Work\label{RWs}}
In this section, we review the related literature and briefly summarize the techniques of the attacks and defenses used in the experiment. Moreover, we introduce GAN structure and its related attack and defense methods as well.

\subsection{Attack Method}
We briefly introduce the classic and lately proposed adversarial attacks, including seven white-box attacks and four black-box attacks, which adopted as baselines in experiments. They are implemented to generate adversarial examples of diverse perturbations, which can fairly verify the defense and detection effect of different methods.

\subsubsection{White-box Attack} White-box attacks mainly use gradient information to determine the direction of perturbation, and increase the value of loss function by adding perturbation until the input is misclassified. 

Goodfellow et al.~\cite{goodfellow2014explaining}  proposed FGSM to find the direction where the gradient of the model changes the most. The algorithm adds perturbation along that direction, which leads to the flip of the predicted label. On the basis of FGSM, Kurakin et al.~\cite{Kurakin2017Adversarial} proposed BIM method, which expanded the operation of increasing the loss function of the classifier to several small steps. As a result, BIM generates smaller perturbation with less transferability. Dong et al. ~\cite{Dong2018Boosting} proposed a kind of momentum based iterative algorithm MI-FGSM, which integrates momentum term into the iterative process of attack to generate more transferable adversarial examples. MI-FGSM does boost the effectiveness of adversarial attacks. projected gradient descent (PGD)~\cite{Madry2018Towards} is an iterative white-box attack. Compared with FGSM of one iteration, it does several iterations with one small step at a time, and each iteration will project the perturbation to the specified range. Consequently it conducts a stronger attack than other previous iterative methods like BIM.  Papernot et al. \cite{papernot2016limitations} proposed Jacobian-based saliency map attack (JSMA) method to calculate the partial derivative of each output of the last layer of neural network to each input feature and find the part which has the greatest impact on the specific output of the classifier. Based on the critical pixels in the saliency map, the algorithm crafts perturbation on the input, generating adversarial examples. Moosavi et al.~\cite{moosavi2016deepfool} generated the minimum norm perturbations by iterative calculation method, namely DeepFool, and gradually pushed the image within the classification boundary to the outside until the wrong predicted label occurred. The "universal" perturbation calculated by Moosavi et al.~\cite{moosavi2017universal} can fool the network on "any" image with high probability, which is called universal adverse perturbations (UAP). Similar to DeepFool, UAP uses the principle of the classification boundary for all images and crafts perturbation invisible to humans. Chen et al.~\cite{chen2021finefool} focused on the object contours and performed FineFool on the basis of attention mechanism. To reduce image distortion caused by large perturbations, Xiao et al.~\cite{xiao2020adversarial} proposed an adaptive gradient-based adversarial attack method named Adaptive Iteration Fast Gradient Method (AI-FGM). By adaptively seeking gradient, attacks are performed with fewer pixel modifications.

\subsubsection{Black-box Attack} 
Black-box attacks do not require the specific parameters of the model; they tend to use the output label or confidence of the model to calculate the perturbation.

LSA~\cite{narodytska2016simple} is a black-box attack based on greedy local search, which adds perturbation to a single pixel or a small part of randomly selected pixels. Adopting the idea of greedy local search, the algorithm constructs a small set of pixels to perturb, improving the effectiveness of the attack. PWA attack~\cite{schott2018towards} starts with an adversarial and performs a binary search between the adversarial and the original for each dimension of the input individually until the input is misclassified. Contrast reduction attack (CRA) reduces the contrast of the input until it is misclassified. Similarly, additive uniform noise attack (AUNA) adds uniform noise to the input, gradually increasing the standard deviation until the model is fooled. In addition, Chen et al.~\cite{chen2019POBAGA} adopted the genetic algorithm and proposed POBA-GA, which achieves white-box comparable attack performances. Wei et al.~\cite{wei2021black} put forward the adversarial attributes and use it for the generation of black-box perturbations, only with the knowledge of predicted probabilities from the model.

Adversarial attacks could be found in the real-world scenario as well. In the field of face recognition~\cite{rozsa2019facial}, autonomous vehicles~\cite{eykholt2018robust} and license plate recognition~\cite{qian2020spot}, malicious manipulation of inputs will cause disastrous results.

\subsection{Defense Method}
Numerous defense strategies have been proposed to deal with adversarial attacks. Based on different purposes, defense against adversarial attacks could be classified into two categories: complete defense and adversarial detector, the former also named re-identification defense.

\subsubsection{Complete Defense}
Complete defense mainly develops in the following three directions: using modified input for training or testing and modifying network parameters or structure, e.g, adding more layers or changing loss function, and using add-on network for defense.

In terms of modified training/input, Goodfellow et al.~\cite{goodfellow2014explaining} and Huang et al.~\cite{huang2015learning} proposed adversarial training, in which adversarial examples were injected into the training set, enhancing the robustness of neural network towards adversarial examples. On this basis, Mummadi et al.~\cite{mummadi2019defending} adopted a new training set to finetune the target model by mixing a large number of adversarial and benign example pairs. Although adversarial training does effectively improve the robustness of the model, it only takes effect on specific attack method with high computation and time cost. Besides, image preprocessing also belongs to defense of input modification. Xie et al.~\cite{xie2017mitigating} found that random resizing and padding of adversarial examples can weaken the strength of the attack. Prakash et al.~\cite{prakash2018deflecting} redistribute the pixel values in adversarial examples by pixel deflection, and then denoise them based on a wavelet, so as to effectively correct class labels. Image preprocessing methods are easy to operate but hard to deal with large perturbed adversarial examples. Zhang et al.~\cite{zhang2021robust} measured the distance between feature distribution of adversarial and benign examples using an optimal transport-based Wasserstein distance. By aligning feature representations, models are no longer easily fooled by a diversity of adversaries.

Network modification mainly provides robustness by changing model structure, loss or activation function. Papernot et al.~\cite{papernot2016distillation} put forward defense distillation, which uses the knowledge of network to shape its own robustness, and proved that it can resist adversarial examples of small perturbation. Guneet S. Dhillon et al.~\cite{dhillon2018stochastic} developed stochastic activation pruning (SAP) method, which prunes the random subset of the activation function and enlarges remaining ones to compensate. As a result, the model gains a certain defense ability against adversarial attacks while maintaining high classification accuracy. Pang et al.~\cite{pang2018towards} minimized the reverse cross entropy in the process of training and proposed RCE, which improves the robustness of the model in the adversarial setting. Most of the network modification defense need to retrain the model, which decreases computational efficiency. Besides, training parameters closely related to defense effect should be carefully chosen to achieve expected results.

The method of using network add-on enables the model to cope with adversarial examples with the help of one or more external models such as autoencoder, GAN or ensemble models. Hlihor et al.~\cite{hlihor2020evaluating} adopted the DAE method to train the autoencoder to minimize the distance between adversarial and benign examples, so as to remove perturbations. Ju et al.~\cite{ju2018relative} studied the neural network ensemble method Ens-D for image recognition task. The ensemble of multiple models can still make robust classification when one of them is hacked. Besides, GAN is introduced to defend against adversarial attacks as well, which will be detailed in Section~\ref{GANbased}.

\subsubsection{Adversarial Detector}
Adversarial detectors can distinguish adversarial examples from benign ones, which serves as an alarming bell in the system. Meng and Chen put forward Magnet~\cite{Meng2017MagNet}, which uses one or more separated detectors and networks to discriminate adversarial examples by approximating the manifold of benign ones. The detection methods designed by Madry~\cite{Madry2018Towards}, Tram{\`e}r~\cite{tramer2018ensemble} and Yin~\cite{Yin2020GAT} are all based on adversarial training, achieving convincing performance in adversarial detection. Metzen et al.~\cite{metzen2017detecting} proposed an extended subnetwork detector to distinguish the real data from adversarial ones, so as to make the network itself more robust. These detection methods need to retrain the subnetwork or classifier, which increases computation burden. Tian et al.~\cite{tian2018detecting} found that adversarial images are sensitive to transformation operations such as rotation and translation, while benign ones are not. They implemented 45 different image transformation methods to detect adversarial examples, achieving a fair good detection ratio. Moreover, another light-weighted detector named ACT-detector~\cite{chen2021act} is designed to achieve better detection rate with much less channels, i.e., 5 channels at least.

\subsection{GAN for Adversarial Attacks and Defenses\label{GANbased}}
GAN, first proposed by Goodfellow et al.~\cite{goodfellow2014generative}, has been widely used in image generation, video prediction, object detection and semantic segmentation. Until now, variant structures have derived from GAN, e.g. WGAN, DCGAN, BEGAN~\cite{gui2020review}, dualgan~\cite{yi2017dualgan} and CoGAN~\cite{liu2016coupled}. In general, GAN is a two-player network structure composed of a generator and a discriminator. The generator is designed to imitate, model and learn the distribution of real data as much as possible, reconstruct the random noise or latent variables, and finally generate realistic examples. The discriminator is to differentiate the real data from generated data. Through the continuous competition between the two internal models, the generation and discrimination ability of them are enhanced until a Nash equilibrium is reached.

In aspect of DNN security, GAN is used to generate malicious examples, posing invisible threats to mission-critical field, and also provides a more powerful way out for defense mechanisms.

For GAN based attacks, Xiao et al.~\cite{xiao2018generating} trained a conditional GAN to implement attacks, namely AdvGAN, which could generate diverse adversarial examples without accessing the targeted model itself. Chen et al.~\cite{chen2020mag} designed MAG-GAN to generate large-scale adversarial examples, and used it as an effective tool to explore the vulnerability and improve the defense capability of DNNs. Liu et al.~\cite{liu2019rob} introduced adversarial examples in the process of training and proposed Rob-GAN, which leads to the enhancement of convergence speed of GAN training and the quality of generated adversarial examples.

On the other hand, Defense-GAN~\cite{samangouei2018defense} and APE-GAN~\cite{jin2019ape} introduced GAN for adversarial defense. As input, the mixture of benign examples and adversarial examples is used to train GAN model until it eliminates adversarial perturbation. The defense of network add-on requires to train the extended network with adversarial examples, and generate the output close to the benign examples through reconstruction. In this way, defensive effect is obtained at the cost of more computation times and lower implementation efficiency.
However, these GAN-based defense methods directly deal with the whole images, increasing the complexity and difficulty of training. Besides, the defense effect cannot be guaranteed  when faced with unknown attacks since they are usually attack-dependent. 
% Inspired by this, we put two GAN structures together in parallel and propose SFE method. The two GANs reconstruct salient and trivial feature in the image respectively. As a result, the output of G approximate to the benign examples when training terminates. These reconstructed features are used for detection and re-identification defense, which can raise the robustness of the model.

\section{Methodology\label{method}}
We first give the definitions of salient feature and trivial features, and then describe SFE in detail. Adversarial example is detected by extraction, separation and comparison of SF and TF. The reconstructed SF of adversarial examples is re-identified by the model to get the correct classification label. Finally, we analyze the convergence to guarantee the stability and robustness of SFE.

\subsection{Preliminary\label{pre}}
A DNN model consists of input, hidden and output layer. For an input example $x$, the hidden layer of DNN gives the output of the feature, denoted as $h(x)$. The result of the output layer is the classification label, denoted as $f_{DNN}(x)$. Here, we adopt the output of the last fully connected layer as feature. 

\textbf{Definition of salient feature (SF) and trivial feature (TF)}.
For an example pair $<X,X*>$, where $x \in X$ and $x* \in X*$ represent benign and its corresponding adversarial example, respectively. The SF and TF of the benign example $x$ are both $h(x)$, where $f_{DNN}(x) \neq{f_{DNN}(x*)}$. For the adversarial example $x*$, the SF and TF will be separated as $h(x)$ and $h(x*)$.

\begin{table}[h]
\centering
\caption{The definition of SF and TF.}
\label{tabel:SFTF}
\resizebox{0.75\linewidth}{!}{
\begin{tabular}{c|c|c|c|c}
\hline
\textbf{example pairs}  & \textbf{input} & \textbf{feature $x_{F}$} & \textbf{salient feature $x_{SF}$} & \textbf{trivial feature $x_{TF}$} \\
\hline
\multirow{2}{*}{$<X,X^*>$} & $x \in X$       & $h(x)$                & $h(x)$                   & $h(x)$   \\\cline{2-5}    
                & $x^* \in X^*$          & $h(x^*)$          & $h(x)$                 & $h(x^*)$ \\
                \hline
\end{tabular}
}
\end{table}

Table~\ref{tabel:SFTF} clearly denotes SF and TF, where SF and TF of the input $x$ are recorded as $x_{SF}$ and $x_{TF}$, and $x_{F}$ denotes the output feature of the hidden layer. By learning the difference between SF and TF, the detector obtains the capability of discriminating adversarial examples. The output label of defense is obtained when the reconstructed SF of adversarial examples is input to the model for re-identification.

\subsection{Framework}
\begin{figure}[h]
\centering
        \includegraphics[width=0.95\linewidth]{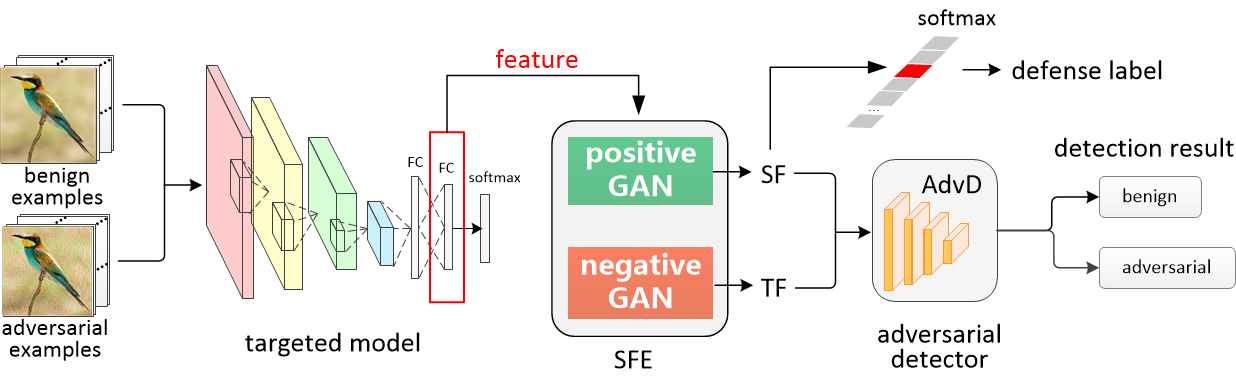}
\caption{The framework of our proposed method, which includes feature extraction, adversarial example detection and re-identification defense.}
	\label{fig:framework}
\end{figure}

The framework of our proposed method is shown in Figure~\ref{fig:framework}, which includes feature extraction and separation of SF and TF, adversarial example detection via training adversarial detector (AdvD) and defense against adversarial examples via re-identification of SF.

Concretely, the targeted model is fed with benign examples and their corresponding adversarial examples at the beginning. The output of the last fully connected layer of the model, considered as high-dimensional feature, is input to SFE. SF and TF are then extracted and separated on the basis of coupled GAN structure. Next, AdvD is trained using the reconstructed SF and TF. Adversarial examples are distinguished by calculating the difference between SF and TF while correct classification labels are obtained by recognition of reconstructed SF, achieving successful defense.

\subsection{Salient Feature Extractor}
\begin{figure}[t]
\centering
        \includegraphics[width=0.8\linewidth]{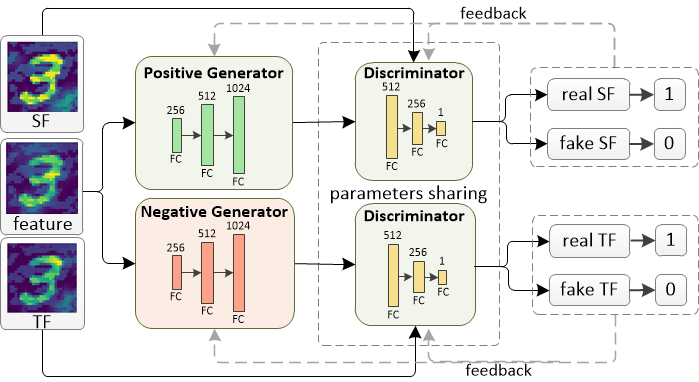}
\caption{The structure of salient feature extractor. SFE consists of coupled GAN structures called positive GAN and negative GAN, which are painted in green and red. Generators and discriminators are all composed of fully connected layers, FC for short, sizes marked above. By reconstructing SF and TF of adversarial examples on the basis of coupled GAN framework, detection and re-identification defense of adversarial examples can be completed at the same time. SF, TF and feature are visualization heatmaps are obtained via Grad-CAM on adversarial examples generated by MI-FGSM on MNIST-CNN1 model. From yellow to blue, the weight that model allocates decreases. }
	\label{fig:SFE}
\end{figure}

SFE consists of two coupled GAN structures called positive GAN and negative GAN, as shown in Fig \ref{fig:SFE}. The former includes a positive generator (PG) and discriminator (D), responsible for learning and generating salient features, while the latter is composed of a negative generator (NG) and D, responsible for trivial features. After inputting the high-dimensional feature of the output of the last fully connected layer of the target model, SFE remaps the features through a coupled GAN structure. Reconstructed SF and TF are provided by PG and NG respectively, which are then input to D for classification. The binary result of D is fed back to the generator and discriminator for optimization of the model parameters. In detail, the output of D is 1 when generated SF and TF are classified as true, vice versa.

The specific structures of generators and discriminators used in the experiment are provided in Table~\ref{tabel:GD}. PG and NG are structurally identical with different functions, so they are trained with different data. To reduce complexity, D of positive GAN and negative GAN share parameters during training with the same parameters. This constraint forces high-level features to decode in the same way in both discriminators, thus better capturing the relationship between SF and TF, separating and generating features similar to that distribution of ground truth. In our experiment, G has the same structure in three datasets with stacked fully connected layers, which are sufficient to deal with high-dimensional image features. The activation function of D on CIFAR-10 and ImageNet datasets is slightly different from that on MNIST. The size of the input layer of G is [$H$,$W$,$C$], the same as that of the image. The size of the output layer of G and that of D is [$H$$\times$$W$$\times$$C$,1], and the output layer of D is [1,1].

\begin{table}[t]
\centering
\caption{The specific structures of G, D and AdvD in our experiment.}
\label{tabel:GD}
\resizebox{1\linewidth}{!}{
\begin{tabular}{cl|c|l|c|lll}
\hline
\multicolumn{2}{c|}{\textbf{Generator}}                                                                                                          & \multicolumn{2}{c|}{\textbf{Discriminator}}                                                                  & \multicolumn{4}{c}{\textbf{Adversarial example   detector}}                                                                               \\ \hline
\multicolumn{1}{l|}{Datasets}                                                                                & structure                        & \multicolumn{1}{l|}{Datasets}                                                 & structure                    & \multicolumn{1}{l|}{Datasets}                                                           & \multicolumn{3}{l}{structure}                    \\ \hline
\multicolumn{1}{c|}{\multirow{13}{*}{\begin{tabular}[c]{@{}c@{}}MNIST \\ CIFAR-10\\  ImageNet\end{tabular}}} & Dense 256                        & \multirow{5}{*}{MNIST}                                                        & Dense 512                    & \multirow{13}{*}{\begin{tabular}[c]{@{}c@{}}MNIST \\ CIFAR-10 \\ ImageNet\end{tabular}} & \multicolumn{3}{l}{Dense 512 activation='ReLU'}  \\
\multicolumn{1}{c|}{}                                                                                        & LeakyReLU(alpha=0.2)             &                                                                               & LeakyReLU(alpha=0.2)         &                                                                                         & \multicolumn{3}{l}{BatchNormalization}           \\
\multicolumn{1}{c|}{}                                                                                        & BatchNormalization(momentum=0.8) &                                                                               & Dense 256                    &                                                                                         & \multicolumn{3}{l}{Dropout 0.25}                 \\
\multicolumn{1}{c|}{}                                                                                        & Dense 512                        &                                                                               & LeakyReLU(alpha=0.2)         &                                                                                         & \multicolumn{3}{l}{Dense 256 activation='ReLU'}  \\
\multicolumn{1}{c|}{}                                                                                        & LeakyReLU(alpha=0.2)             &                                                                               & Dense 1 activation='tanh'    &                                                                                         & \multicolumn{3}{l}{BatchNormalization}           \\ \cline{3-4}
\multicolumn{1}{c|}{}                                                                                        & BatchNormalization(momentum=0.8) & \multirow{8}{*}{\begin{tabular}[c]{@{}c@{}}CIFAR-10 \\ ImageNet\end{tabular}} & Dense 512                    &                                                                                         & \multicolumn{3}{l}{Dropout 0.25}                 \\
\multicolumn{1}{c|}{}                                                                                        & Dense 1024                       &                                                                               & LeakyReLU(alpha=0.2)         &                                                                                         & \multicolumn{3}{l}{Dense 128 activation='ReLU'}  \\
\multicolumn{1}{c|}{}                                                                                        & LeakyReLU(alpha=0.2)             &                                                                               & Dense 256                    &                                                                                         & \multicolumn{3}{l}{BatchNormalization}           \\
\multicolumn{1}{c|}{}                                                                                        & BatchNormalization(momentum=0.8) &                                                                               & LeakyReLU(alpha=0.2)         &                                                                                         & \multicolumn{3}{l}{Dropout 0.25}                 \\
\multicolumn{1}{c|}{}                                                                                        & Dense activation='tanh'          &                                                                               & Dense 1 activation='sigmoid' &                                                                                         & \multicolumn{3}{l}{Dense 64 activation='ReLU'}   \\
\multicolumn{1}{c|}{}                                                                                        &                                  &                                                                               &                              &                                                                                         & \multicolumn{3}{l}{BatchNormalization}           \\
\multicolumn{1}{c|}{}                                                                                        &                                  &                                                                               &                              &                                                                                         & \multicolumn{3}{l}{Dropout 0.125}                \\
\multicolumn{1}{c|}{}                                                                                        &                                  &                                                                               &                              &                                                                                         & \multicolumn{3}{l}{Dense 1 activation='sigmoid'} \\ \hline
\end{tabular}
}
\end{table}

The training process of SFE is shown in Algorithm 1. Benign and adversarial examples will be input into the targeted model, and the output of the last fully connected layer will be taken out as high-dimensional feature for training SFE.

We use mean square error(MSE) as the optimization objective to minimize the distance between the input features and the corresponding generated features, approximating the generated data to real data. The parameters of PG and D are updated alternately during the training process, and loss function is defined as:
\begin{align}
loss_{PG}&=MSE(\rm PG(x_F),x_{SF})+CE(D(PG(x_F)),1)\\
loss_{ D_{PG}}&=CE(\rm D(PG(x_F)),0)+CE(D(x_{SF}),1)
\label{equ:lossPG}
\end{align}
where $MSE(\cdot,\cdot)$ denotes mean square error, $CE(\cdot,\cdot)$denotes cross-entropy of binary classification. $PG(\cdot)$ and $D(\cdot)$ represents the output of the generator and discriminator in positive GAN respectively. $x_F$ denotes feature output by the last fully connected layers, while $x_{SF}$ denotes salient feature.

Similarly, when training negative GAN, the parameters of NG and D are updated alternately, and the loss functions are defined as follows:
\begin{align}
loss_{NG}&=MSE(\rm NG(x_F),x_{TF})+CE(D(NG(x_F)),1)\\
loss_{ D_{NG}}&=CE(\rm D(NG(x_F)),0)+CE(D(x_{TF}),1)
\label{equ:lossNG}
\end{align}
where $NG(\cdot)$ and $D(\cdot)$ denotes output of generator and discriminator in negative GAN separately. $x_{TF}$ denotes trivial feature.

The calculation formula of the average square error is defined as follows:
\begin{equation}
    MSE=\frac{1}{n}\sum_{i=1}^{n}(y_i-\hat{y}_i)^2
\end{equation}
where $y_i$ represents the true class label and $\hat{y}_i$ denotes the predicted value of the model.

Cross-entropy of binary classification is calculated as follows:
\begin{equation}
    CE=-(y\cdot{log(\hat{y})}+(1-y)\cdot{log(1-\hat{y}}))
\end{equation}
where $\hat{y}$ is the probability of correct prediction while $y$ represents ground truth label. Concretely, the value is 1 if the example is positive, vice versa.

The pseudo-code of SFE is presented in Algorithm 1.

\begin{tabular}{p{0.01\linewidth}p{0.9\linewidth}}
\toprule
  \hline
   & \textbf{Algorithm 1}: The training process of SFE \\
  \hline
   & \textbf{Input}: Benign examples $X=\{x_1,x_2,…,x_m\}$. Adversarial examples $X*=\{x_1^*,x_2^*,…,x_m^*\}$. High-dimensional feature of the model $x_F=H(X)=\{h(x_1),h(x_2),…,h(x_m)\}$. $x_F^*=H(X^*)=\{h(x_1^*),h(x_2^*),…,h(x_m^*)\}$. The minibatch size $m_b$. Input parameters, the number of steps to pre-train the discriminator $k_D$.  \\
   & \textbf{Output}: salient feature $x_{SF}$ and trivial feature $x_{TF}$.   \\
  \hline
  1. & \textbf{for} $k_D$ steps \textbf{do} \\
  2. & \quad Sample minibatch of $m$ benign feature from $x_F$.\\
  3. & \quad Sample minibatch of $m$ adversarial feature from $x_F^*$.\\
  4. & \quad Obtaining generated data $\{x_{SF}^{\sim 1},x_{SF}^{\sim 2},…,x_{SF}^{\sim m}\}$, $x_{SF}^{\sim i}=\rm PG(x_F^*)$.\\
  5. & \quad Obtaining generated data $\{x_{TF}^{\sim 1},x_{TF}^{\sim 2},…,x_{TF}^{\sim m}\}$, $x_{TF}^{\sim i}=\rm NG(x_F^*)$.\\
  6. & \quad Update the discriminator parameter $\theta_D$ to minimize $\{loss_{D(PG)}+loss_{D(NG)}\}$.\\
  7. & \textbf{end for}  \\
  8. & \textbf{for} number of training iterations \textbf{do} \\
  9. & \quad Sample new minibatch of $m$ adversarial feature  from $x_F^*$.\\
  10. & \quad Update PG parameter $\theta_{\rm PG}$ to minimize $\{loss_{\rm PG}\}$.\\
  11. & \quad Update NG parameter $\theta_{\rm PG}$ to minimize $\{loss_{\rm NG}\}$.\\
  12. & \quad Update the discriminator parameter $\theta_D$ to minimize $\{loss_{D(\rm PG)}+loss_{D(\rm NG)}\}$.\\
  13. & \textbf{end} \\
  14. & The minimization can use any standard optimization learning rule. We used Adam optimizer in our experiments.\\
  \hline
  \bottomrule
\end{tabular}

\subsection{Adversarial Examples Detection via AdvD}
By using coupled GAN structure, we can separate and extract SF and TF in the input image. As defined in Section~\ref{pre}, SF in benign examples is similar to TF, while there is a difference between SF and TF in adversarial examples. Based on that definition, we design the adversarial example detector (AdvD) to determine whether the input data is adversarial.

 AdvD consists of five fully connected layers, whose specific structure is shown in Table~\ref{tabel:GD}. The input layer size of AdvD is [$H$$\times$$W$$\times$$C$,1], the same as the output of the generator in SFE. And the size of its output layer is [1,1].
% \begin{figure}[ht]
% \centering
%         \includegraphics[width=0.8\linewidth]{detection.png}
% \caption{The framework of adversarial examples detection.
% }
% 	\label{fig:detection}
% \end{figure}
% The process of detection against adversarial examples is shown in Fig.~\ref{fig:detection}.

 As Figure~\ref{fig:framework} shows, after the training of SFE finishes, its output will be used to train AdvD.  The output of PG, the SF of benign and adversarial examples, and the output of NG, TF, will be concatenated to generate the training set of AdvD. During training, the input of AdvD is the concatenated feature, and the output is the detection result. "benign" means benign example, marked with 0, and "adversarial" means adversarial example, marked with 1. During detection, The loss function of AdvD is defined as follows:
\begin{equation}
\begin{aligned}
    loss_{\rm AdvD}=&CE(\rm AdvD(Concat(\rm PG(h(x)),NG(h(x)))),0)+\\
    &CE(\rm AdvD(Concat(\rm PG(h(x^*)),NG(h(x^*)))),1)
    \end{aligned}
\end{equation}
where $x$ denotes benign examples while $x^*$ denotes adversarial examples. $\rm AdvD(\cdot)$  represents the output of AdvD. $Concat(\cdot)$ is concatenate function, which does concatenate operation on the last dimension of the matrix. PG and NG represent generators in positive GAN and negative GAN respectively, whose inputs are $h(x)$, the high-dimensional feature of the last fully connected layer of the targeted model.

For the well-trained SFE, The parameters of targeted model, PG, and NG are fixed and the parameters of AdvD are updated by $min$  ${loss_{AdvD}}$. After training, AdvD will give detection results when the mixture of benign and adversarial examples is input to it. 

\subsection{Adversarial Examples Re-identification via SF}
As defined in Section~\ref{pre}, SF of the adversarial examples are the same as that of its corresponding benign examples and the hidden layer of the model. In the original model, the high-dimensional image features contain important information closely related to the label, which are fed into the next layer for image classification. Positive GAN in SFE reconstructs the high-dimensional feature and strengthens that important features. Therefore, for well-trained SFE, SF, output from PG, still retains critical information for classification, which can be adopted to correct labels of adversarial examples. Similar to the detection process, we only use generators for the well-trained SFE model. Correct classification results will be given when reconstructed SF is input to the targeted model.

\subsection{Convergence Analysis}
The optimization objective function of GAN is a minimax game corresponding to two players, generator G and discriminator D. To prove the convergence of SFE, we first consider the optimal of any given G. Take PG as an example for the following mathematical proof.

\textbf{Proposition 1}: For a fixed PG, D is optimized by PG, that is $\exists \rm D(x)^*=\frac{P_{data}(x)}{P_{data(x)}+P_{\rm PG}(x)}$.

\textbf{Proof}: Given PG, D is maximized for 
\begin{equation}
\begin{aligned}
    V(PG,D)=&E_{x\sim P_{data}}[log\rm D(x)]+E_{x\sim P_{PG}}[log(1-D(x))]\\
    =&\int_{x}P_{data}(x)log\rm D(x)dx+\int_{x}P_{PG}(x)log(1-D(x))dx\\
    =&\int_{x}[P_{data}(x)log\rm D(x)+P_{PG}(x)log(1-D(x))]dx
    \end{aligned}
\end{equation}
This formula can be simplified as $f(\rm D)=alog(D)+blog(1-D)$. If and only if $P_{\rm PG}(x)=P_{SF}(x)$, D reaches the optimal value $\rm D(x)^*=\frac{P_{data}(x)}{P_{data(x)}+P_{PG}(x)}$. The same procedure may be easily adapted to obtain the optimum result of D in NG.

\textbf{Theorem 1}: if and only if $P_G=P_{data}$, the global minimum of training criterion $C (\rm G)$ = max $V (\rm G, D)$ can be reached.

\textbf{Proof}: According to GAN theory, if and only if $P_G=P_{data}$. At that point, $V (\rm G, D)$ achieves the value $-log 4$. For any G, we can substitute the optimal discriminator $D^*$ obtained in the previous step into $C (\rm G)$ = max $V (\rm G, D)$ and obtain that $C(\rm G)=-log4+2JS(P_{data}\vert P_G)$ where $JS(\cdot)$ denotes Jensen-Shannon divergence. $P_G=P_{data}$ is the possible value of $C (\rm G)$. This concludes the proof.

If PG, NG and D have enough capacity, D can reach the optimal value of given PG and NG.

\section{Experiments and Analysis\label{Exp}}
To illustrate general significance of the results, extensive experiments have been carried out to testify the state-of-the-art performance of SFE, including the following parts: 
\begin{itemize}
    \item \textbf{RQ1}: Is SFE capable of shielding the model from adversarial manipulation of inputs? And is it competitive enough when compared with baselines about detection and defense results?
    \item \textbf{RQ2}: Does SFE have good transferability over attacks?
    \item \textbf{RQ3}: Can SFE meet the need of efficiency with low time complexity?
    \item \textbf{RQ4}: Can SFE provide a visual understanding of interpretable defense?
\end{itemize}

\subsection{Setup\label{setup}}
\textbf{Platform}:
i7-7700K 4.20GHzx8 (CPU),
TITAN Xp 12GiB x2 (GPU),
16GBx4 memory (DDR4),
Ubuntu 16.04 (OS),
Python 3.6,
Tensorflow-gpu-1.3,
Tflearn-0.3.2.~\footnote{Tflearn can be downloaded at \emph{https://github.com/tflearn/tflearn.}}.

\textbf{Datasets}: We verify the effectiveness of the proposed method on the
MNIST~\footnote{MNIST can be download at \emph{http://yann.lecun.com/exdb/mnist/}},
CIFAR10\footnote{CIFAR10 can be download at \emph{ https://www.cs.toronto.edu/~kriz/cifar.html}}, and
ImageNet\footnote{ImageNet can be download at \emph{http://www.image-net.org/}} datasets.
MNIST includes 60000 training examples and 10000 testing examples. Each of it is a 28$\times$28 pixel gray handwritten digital image, marked with ten classes range from 0 to 9. CIFAR-10 dataset consists of 60000 32$\times$32 color images of 10 classes such as airplane, bird, ship and frog, with 6000 in each class. ImageNet project is a large visual database for visual object recognition research. It contains more than 14 million images, covering up to 20000 categories. In our experiment, we selected 10 classes with a total of 13500 images for detection and defense test.

\textbf{DNNs}:
A Variety of different models are adopted in our experiments. For MNIST, we crafted two self-trained ConvNets, whose network structure is shown in Table~\ref{tabel:mnist_model}. For CIFAR-10, we trained AlexNet~\cite{krizhevsky2012imagenet} and VGG19~\cite{simonyan2014very}, whose classification rate are 99.82$\%$ and 99.88$\%$ respectively. Besides, Inception v3 (Inc-v3)~\cite{szegedy2016rethinking} and VGG19 are used in the experiment of ImageNet dataset. Their recognition accuracy go to 99.48$\%$ and 99.50$\%$.

\begin{table}[t]
\centering
\caption{The network structure for MNIST.}
\label{tabel:mnist_model}
\resizebox{0.5\linewidth}{!}{
\begin{tabular}{lll}
 \hline
\textbf{Layer Types}              & \textbf{CNN1} (acc=99.79\%) & \textbf{CNN2} (acc=99.76\%) \\ \hline
Conv+ReLU              & 5*5*32                     & 5*5*16                     \\
Max Pooling            & -                          & 2*2                        \\
Conv+ReLU              & 5*5*64                     & 5*5*32                     \\
Max Pooling            & 2*2                        & 2*2                        \\
Flatten                & 1                          & -                          \\
Dropout                & 0.5                        & 0.25                       \\
Flatten                & -                          & 1                          \\
Dense                  & 128                        & 128                        \\
Dropout                & 0.5                        & 0.5                        \\
Softmax                & 10                         & 10                         \\ \hline
\end{tabular}
}
\end{table}

\textbf{Attack methods}:
We used eleven attack methods, which adopt various algorithms to generate adversarial examples of different perturbation size and distribution, so as to prove the effectiveness of defense and detection of SFE. The white-box attacks include FGSM~\cite{goodfellow2014explaining}, DeepFool (L2)~\cite{moosavi2016deepfool}, PGD~\cite{Madry2018Towards}, MI-FGSM~\cite{Dong2018Boosting}, BIM~\cite{Kurakin2017Adversarial}, JSMA~\cite{papernot2016limitations} and UAP~\cite{moosavi2017universal} while black-box attacks contain CRA~\cite{FoolboxCRA}, AUNA~\cite{FoolboxAUNA}, PWA~\cite{schott2018towards} and LSA~\cite{narodytska2016simple}. Attack success rate and average perturbation sizes are shown in Table~\ref{tab:attackasr} and Table~\ref{tab:attackpara}, where $\rho_{adv}$ denotes the average perturbation size of each pixel after normalization to $[0,1]$.

\begin{table}[ht]
\caption{The attack success rate of adversarial examples. The pixel values of each image are normalized to [0,1]. The data in the table represents the attack success rate.}
\label{tab:attackasr}
\resizebox{\linewidth}{!}{
\begin{tabular}{ccccccccccccc}
\hline
\multirow{2}{*}{\textbf{Datasets}} & \multirow{2}{*}{\textbf{Models}} & \multicolumn{11}{c}{\textbf{Attack Methods}}                                                                        \\\cline{3-13}
                                   &                                  & DeepFool & FGSM    & PGD      & UAP      & BIM      & MI-FGSM & JSMA     & PWA      & CRA      & AUNA     & LSA     \\
                                   \cline{1-13}
\multirow{2}{*}{MNIST}             & CNN1                             & 86.71\%  & 84.16\% & 91.44\%  & 90.07\%  & 99.99\%  & 99.99\% & 99.93\%  & 99.82\%  & 91.37\%  & 90.72\%  & 99.26\% \\
                                   & CNN2                             & 90.96\%  & 86.79\% & 92.41\%  & 89.76\%  & 100.00\% & 99.98\% & 99.98\%  & 99.87\%  & 89.87\%  & 96.23\%  & 95.18\% \\
                                   \cline{2-13}
\multirow{2}{*}{CIFAR-10}          & AlexNet                          & 88.18\%  & 89.03\% & 86.71\%  & 89.83\%  & 96.22\%  & 89.57\% & 99.85\%  & 87.45\%  & 92.95\%  & 99.99\%  & 95.37\% \\
                                   & VGG19                            & 96.66\%  & 88.29\% & 99.95\%  & 88.82\%  & 99.87\%  & 99.76\% & 100.00\% & 97.60\%  & 91.67\%  & 100.00\% & 90.03\% \\
                                   \cline{2-13}
\multirow{2}{*}{ImageNet}          & Inc-v3                           & 89.80\%  & 89.70\% & 95.03\%  & 99.15\%  & 98.58\%  & 96.70\% & 99.25\%  & 100.00\% & 94.50\%  & 99.60\%  & 94.50\% \\
                                   & VGG19                            & 98.56\%  & 89.66\% & 100.00\% & 100.00\% & 99.75\%  & 99.50\% & 99.75\%  & 96.15\%  & 100.00\% & 99.50\%  & 90.00\% 
                                   \\
                                   \hline
\end{tabular}
}
\end{table}

\begin{table}[t]
    \caption{The average perturbation size and attack parameters of adversarial examples.}
    \label{tab:attackpara}
    \resizebox{0.95\linewidth}{!}{
    \begin{tabular}{lcl}
        \hline
\textbf{Attack} & \textbf{$\rho_{adv}$} & \textbf{Parameters}                                       \\\hline
DeepFool        & 0.002          & epsilon=1e-6,maxiter=100                                                             \\
FGSM            & 0.008          & epsilon=0.3,stepsize=0.05,iterations=10.                                             \\
PGD             & 0.001          & epsilon=0.3,stepsize=0.01,iterations=100                                             \\
UAP             & 0.002          & delta=0.2,maxiter=20                                                                 \\
BIM             & 0.001          & epsilon=0.3,stepsize=0.05,iterations=10,                                             \\
MI-FGSM         & 0.001          & epsilon=0.3,stepsize=0.06,iterations=10,decayfactor=1                                \\
JSMA            & 0.002          & maxiter=2000,num random targets=0,fast=True,theta=0.1, max perturbations per pixel=7 \\
PWA             & 0.006          &                                                                                      \\
CRA             & 0.031          & epsilons=1000                                                                        \\
AUNA            & 0.012          &                                                                                      \\
LSA             & 0.061          & perturbation parameter r=1.5,p=10.0,d=5,max perturbations per pixel=5,maxiter=150  \\
\hline
    \end{tabular}
}
\end{table}

\textbf{Defense baselines}:
Nine defense methods and five detection baselines were applied in our experiment. Defense methods includes resize(RS), rotate(RT), random resize(RRS),random rotate(RRT)~\cite{xie2017mitigating}, RCE~\cite{pang2018towards}, Ens-D~\cite{ju2018relative}, DAE~\cite{hlihor2020evaluating}, Defense-GAN~\cite{samangouei2018defense} and APE-GAN~\cite{jin2019ape}. The settings of different defenses are detailed in Table~\ref{tab:defensepara}. As for detection, 45C-Detector~\cite{tian2018detecting},Perturbation detection (Per-D)~\cite{metzen2017detecting}, adversarial training-based detection (AdvT-D)~\cite{tramer2018ensemble}, PGD-based detection (PGD-D) ~\cite{Madry2018Towards} and GAT~\cite{Yin2020GAT} are used to make fair comparison with SFE.

\begin{table}
\caption{The operation setting of different defenses.}
    \label{tab:defensepara}
    \resizebox{0.95\linewidth}{!}{
\begin{tabular}{ll}
\hline
\textbf{Defense}    & \textbf{Operation}                                                            \\\hline
resize (RS)         & $(H,W)\to (H/2,W/2)\to (H,W)$                                                         \\
random resize (RRS) & $(H,W)\to (H',W')\to (H,W)$, where $H'\in[H/2,H]$, $W'\in[W/2,W]$                               \\
rotate (RT)         & rotate -$45^{o}$ $\to$ rotate $45^{o}$, fill in missing values caused by rotation             \\
random rotate (RRT) & rotate rotate -$r$ $\to$ rotate $r$, fill in missing values caused by rotation, where $r\in[0^{o},45^{o}]$ \\
\hline
\end{tabular}
}
\end{table}

\textbf{Metrics}:
The metrics used in the experiments are defined as follows:

\textcircled{1} Classification accuracy: $acc=\frac{n_{true}}{N}$, where $n_{true}$is the number of clean examples correctly classified by the targeted model and $N$ denotes the total number of benign images.

\textcircled{2} Attack success rate: $ASR = \frac{N_{adv}}{N}$, where $N_{adv}$ denotes the number of adversarial examples misclassified by the targeted model after attacks.

\textcircled{3} Perturbation L2-norm: $\rho_{adv}=||x_{adv}-x||_2$, where $x$ and $x_{adv}$ are benign and its corresponding adversarial example respectively, and $||\cdot||_{2}$ represents L2 norm.

\textcircled{4} Detection rate, $DR = \frac{N_{det}}{N}$,where $N_{det}$ is the number of input examples that are detected by model.

\textcircled{5} Defense success rate, $DSR = \frac{N_{def}}{N_{adv}}$,where $N_{def}$ denotes the number of adversarial examples correctly classified by the target model after defense.

\textcircled{6} Within-class distance, $FSA_k = 1-\frac{norm(\sum{j=1}^{n_k}dist(f_{x_j,k},f_{ck}))}{n_k}$,$FSA=\frac{\sum{i=1}^Kd^2(k)}{K}$~\cite{chen2020roby}, where $n_k$ denotes the number of examples belonging to class $k$. $f_{x_j,k}$ denotes the feature vector of the example $x_j$ belonging to the $k-th$ class in the high-dimensional feature space. $f_{ck}$ represents the center of class $k$ in the feature subspace. $K$ denotes the number of total class and $dist(\cdot)$ represents the normalization function. For each class, we sum $FSA_k$ and average it to gain the final $FSA$. The model with smaller with-in class distance always has higher classification accuracy.

\textcircled{7} Among-class distance, $FSD_{k,k+1} =dist(f_{ck},f_{ck+1})$,$FSD=\frac{1}{K(K+1)/2}\sum\limits_{i=1}^{K-1} \sum\limits_{j=i+1}^KFSD_{i,j}$~\cite{chen2020roby}, where $f_{ck}$ and $f_{ck+1}$ represent the center of class $k$ and $k+1$. The final $FSD$ is calculated by averaging the sum of $FSD_{i,j}$ in each class. Similar to with-in class distance, the model with larger among-class distance performs better in identification and recognition.

\subsection{Comparison of Detection Results\label{detection}}
\begin{figure}[htbp]
  \centering
  \subfigure[MNIST-CNN1]{
  \includegraphics[width=0.48\linewidth]{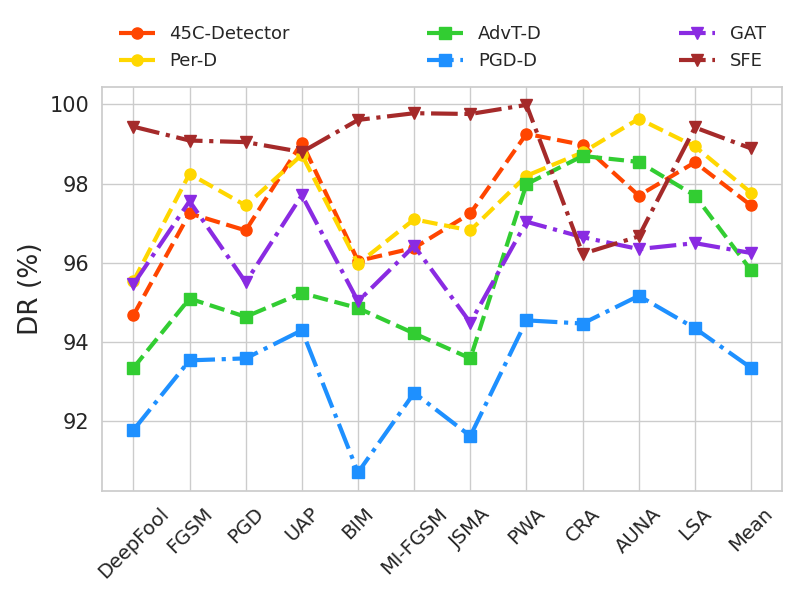} }
  \subfigure[MNIST-CNN2]{
  \includegraphics[width=0.48\linewidth]{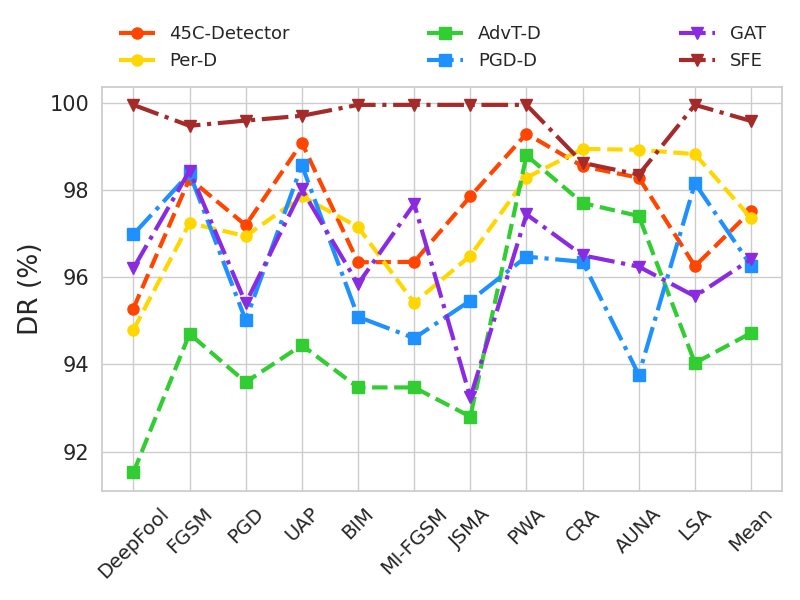} }\\
  \subfigure[CIFAR-10 AlexNet]{
  \includegraphics[width=0.48\linewidth]{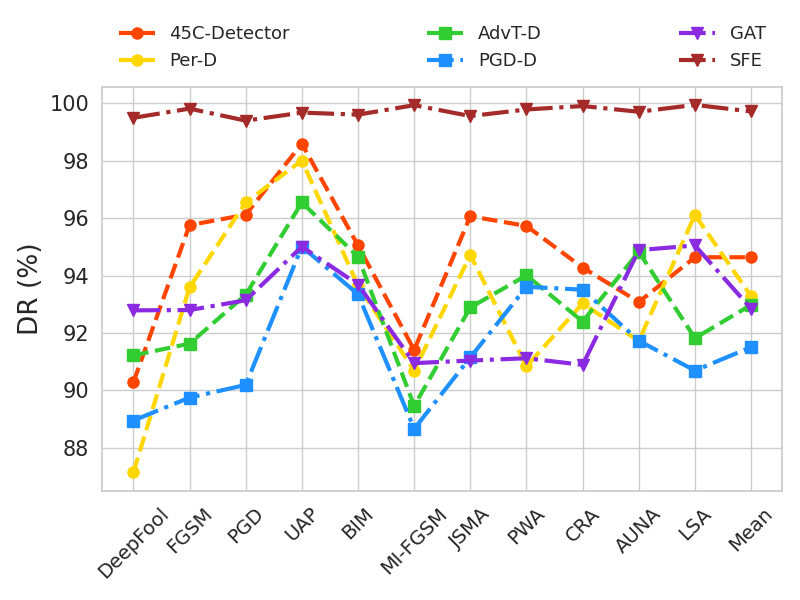} }
  \subfigure[CIFAR-10 VGG19]{
  \includegraphics[width=0.48\linewidth]{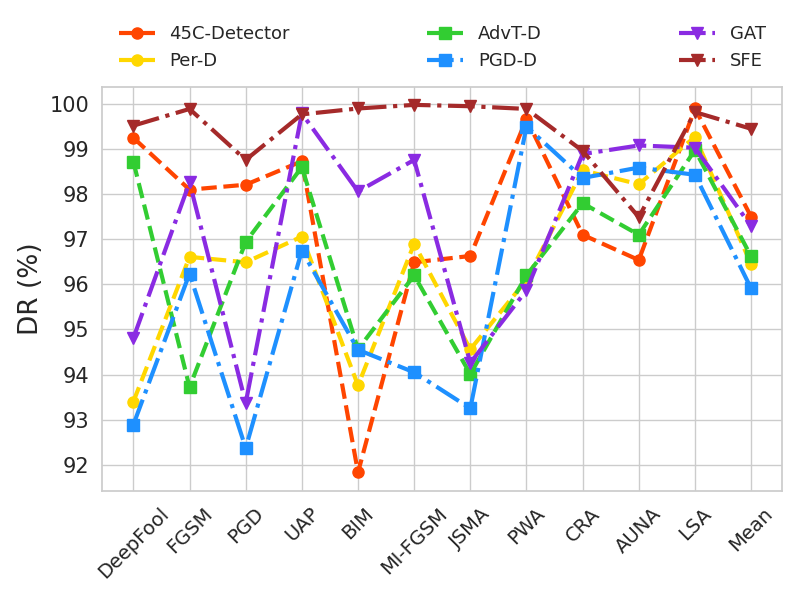} } \\
  \subfigure[ImageNet Inc-v3]{
  \includegraphics[width=0.48\linewidth]{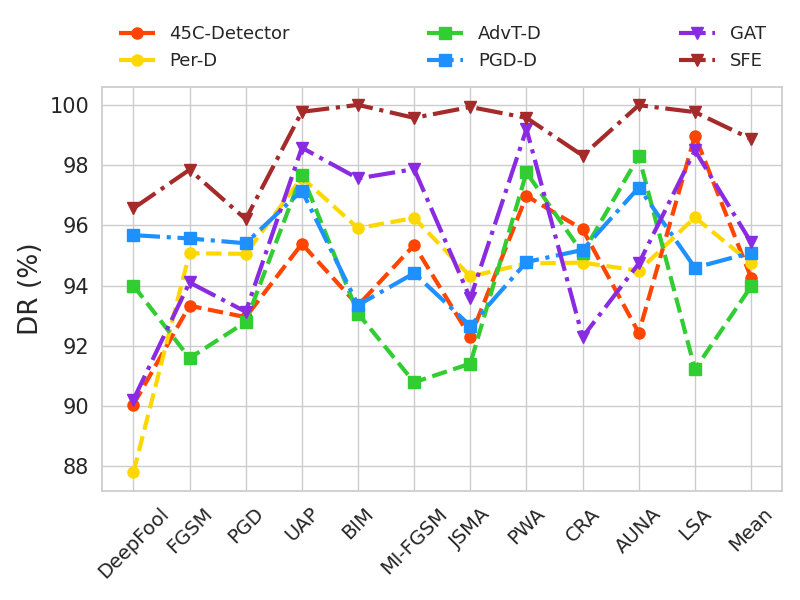} }
  \subfigure[ImageNet VGG19]{
  \includegraphics[width=0.48\linewidth]{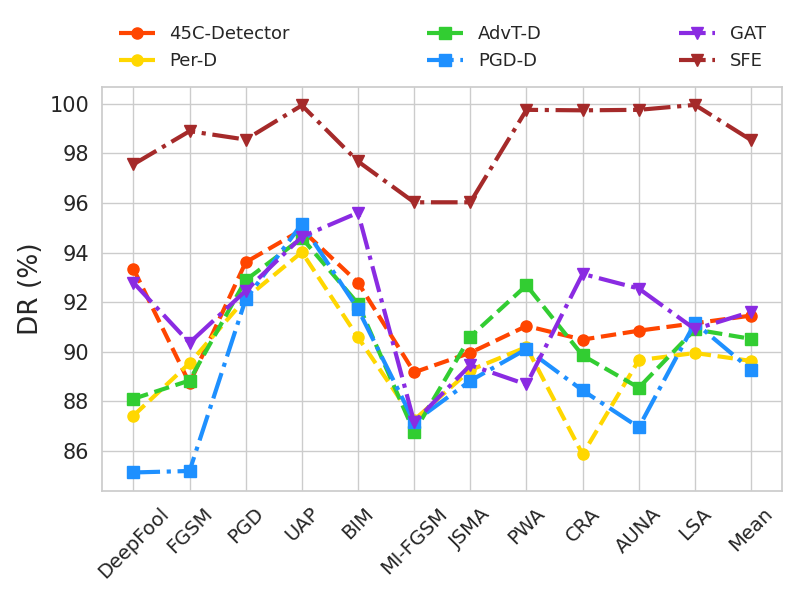} }\\
  \caption{Comparison of detection results against various adversarial attacks.}
  \label{fig:exp_detection}
\end{figure}

In this section, we will focus on the aspect of detection in question \textbf{RQ1}, and use detection rate (DR) to measure the detection effect of SFE on adversarial examples. Comparison will be made with SFE and detection baselines. Fig.~\ref{fig:exp_detection} shows the detection results of SFE and its baselines on three different datasets and six models, where the ordinate represents DR and the last column mean denotes the average of it. In the experiment, we mix the randomly selected adversarial and benign examples then input them to the model for detection. The image of training and testing set is 7:3.

According to the experiment results shown in Fig.~\ref{fig:exp_detection}, SFE has high detection accuracy for a variety of datasets and models, and is superior to five baselines in most cases. Reaching almost 100\% on MNIST and CIFAR-10, DR of SFE slightly decreased on ImageNet dataset. The reason goes to the fact that image size on the ImageNet is larger, so the difference between the salient and trivial features after reconstruction of SFE are obscure, which somewhat reduces the detection performance. As indicated in Fig.~\ref{fig:exp_detection}, the detection effects of SFE remain stable. Meanwhile, DR of black-box attacks are practically as good as that of white-box attacks, which indicates that SFE can accurately distinguish differences between features for comparison and detection. In this way, a good transferability between various attacks is guaranteed. In contrast to other detection algorithms, their DR fluctuates greatly among different attacks. This can be attributed to the fact that there exist differences in the image transformation and training of the adversarial examples generated by various attacks, leading to unstable and unreliable performance. However, SFE shows its vulnerability to attacks with large perturbation, such as CRA and LSA, and has a decline in DR. This is because large perturbation destroys important features in some cases, so the reconstructed SF and TF can not fully reflect the classification information of the original image.

\subsection{Comparison of Defense Results\label{defense}}
In the previous section, we have discussed the detection capability of SFE against adversarial examples. And in this section, we conduct the defense experiment to answer the defense aspects of \textbf{RQ1} by using DSR to measure the effect of re-identification defense.

\begin{figure}[htbp]
  \centering
  \subfigure[MNIST-CNN1]{
  \includegraphics[width=0.48\linewidth]{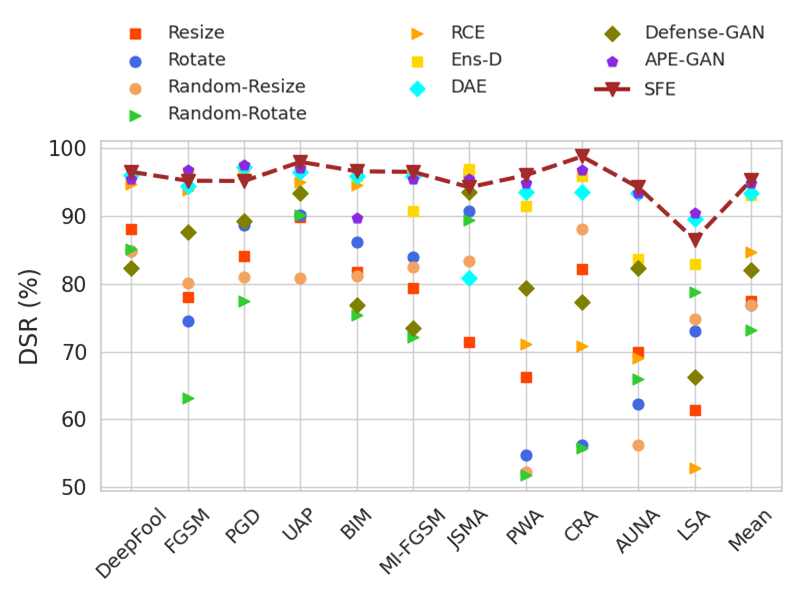} }
  \subfigure[MNIST-CNN2]{
  \includegraphics[width=0.48\linewidth]{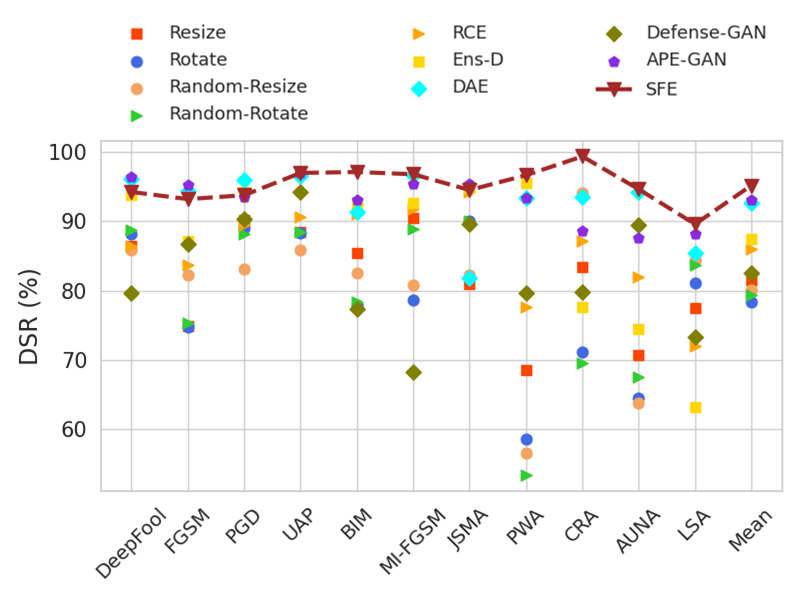} }\\
  \subfigure[CIFAR-10 AlexNet]{
  \includegraphics[width=0.47\linewidth]{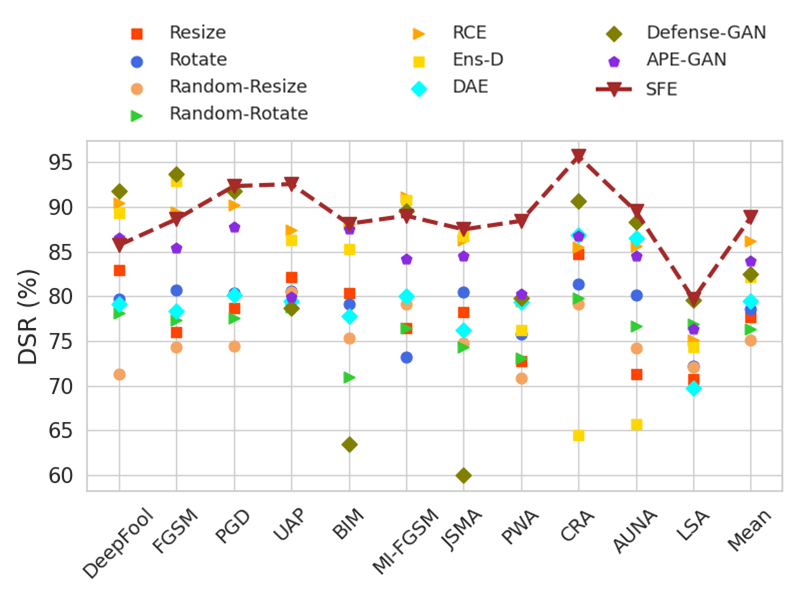} }
  \subfigure[CIFAR-10 VGG19]{
  \includegraphics[width=0.47\linewidth]{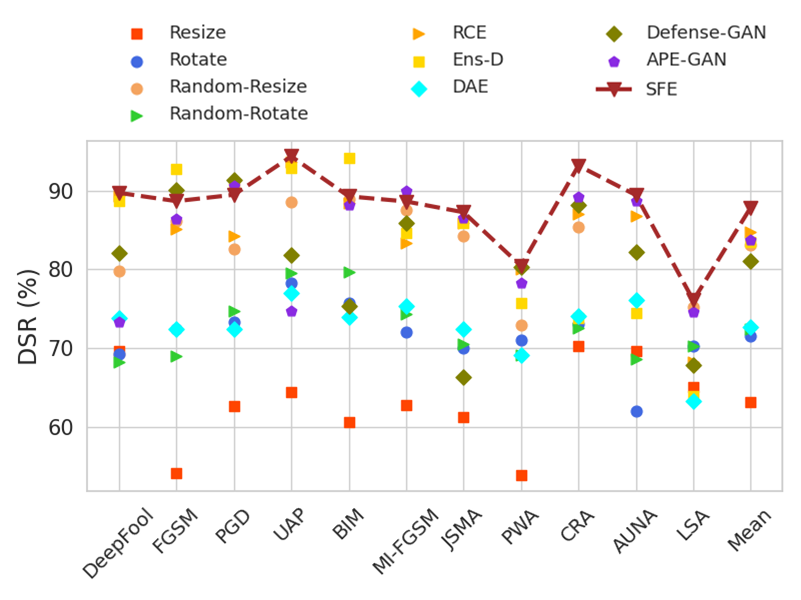} } \\
  \subfigure[ImageNet Inc-v3]{
  \includegraphics[width=0.47\linewidth]{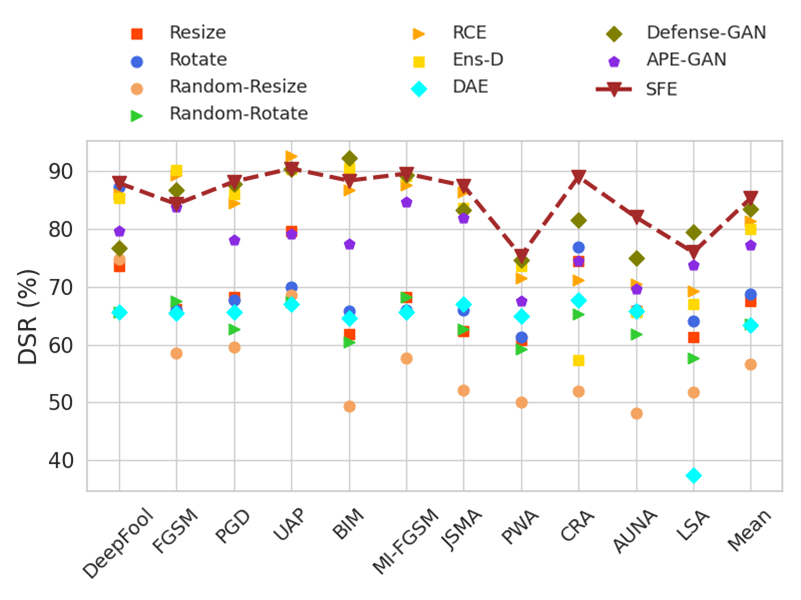} }
  \subfigure[ImageNet VGG19]{
  \includegraphics[width=0.47\linewidth]{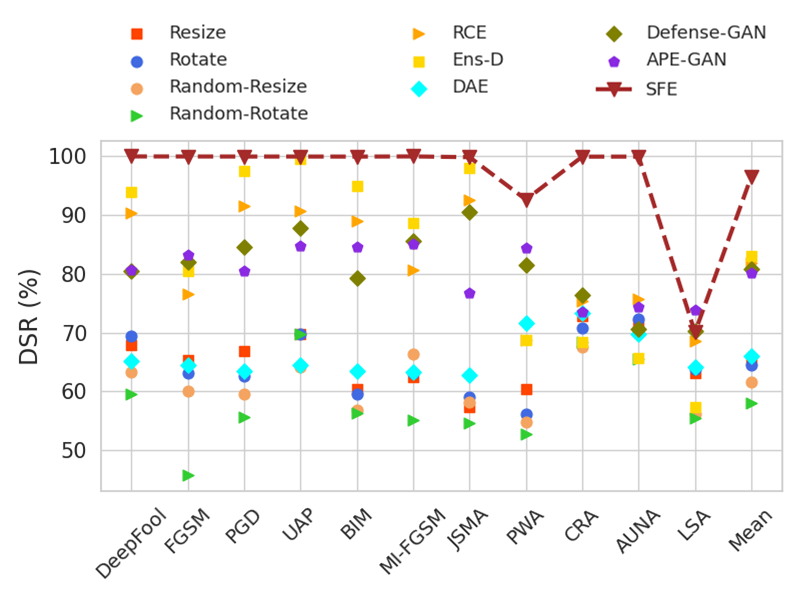} }\\
  \caption{Comparison of defense results for different classifiers on different datasets.}
  \label{fig:exp_defense}
\end{figure}

Fig.~\ref{fig:exp_defense} shows the experimental results of SFE and nine other defense methods on different datasets, models, and attacks, where the ordinate represents DSR, and the last column is the average of it. The results of SFE are denoted by broken brown lines while other defense methods are represented by scattered points with different shapes.

Generally speaking, our proposed method significantly enhances the robustness of the model against adversarial attacks. It can be clearly observed that the average DSR on three datasets is about 90\%, which reaches the highest among all baselines. This indicates that SFE can successfully reconstruct the salient features of different kinds of adversarial examples, making them similar to benign images. Therefore, when salient features are input into the model for re-identification, the correct classification results can be obtained. With regard to different attacks, SFE has a certain defense ability against black-box attacks, but the effect is inferior to white box, opposite to detection results. One possible reason is that black-box attacks tend to have larger perturbation, which may pose a negative impact in the process of extracting and reconstructing salient features. Similarly, for white-box attacks like FGSM with large perturbation, the effect of SFE on feature extraction and reconstruction is slightly reduced, which leads to the decline of DSR. As for different datasets, the defense capability of SFE is affected by data distribution as well. For large-scale images and complex data sets such as ImageNet, the difficulty of feature extraction increases, which affects the defense effect to a certain extent. However, with the increase of the complexity of data sets, DSR of SFE remains around 85\%. In comparison with baseline, we can find that image transformation operations such as resize and rotate do weaken the attack strength, but only when the parameters are selected properly can it get satisfactory defense effect. Although defense results of RCE and Ens-D are close to SFE in DSR, SFE outperforms them in most situations, as well as APE-GAN and defense-GAN.
\begin{framed}
    For \textbf{RQ1}, we can conclude that SFE has quite competitive performance in both detection and re-identification defense. It is slightly superior when compared with baselines. When a malicious perturbation is input to the model, SFE can play a good alarming role while correct classification labels for most of the adversarial examples.
\end{framed}

\subsection{Analysis of Within and Among-class Distance}
In order to better verify the effectiveness of SFE, we pay attention to the feature manifold during the process, and further calculate the within-class and among-class distance of benign example, its corresponding adversarial example and the defense model output. As shown in the formula in Section~\ref{setup}, within-class distance with a smaller value indicates a more concentrated feature distribution of the same class in the image. Similarly, among-class distance with larger value shows that the feature distribution of different classes is more dispersed, which is easy for the model to distinguish.

\begin{figure}[t]
    \centering
    \includegraphics[width=0.8\linewidth]{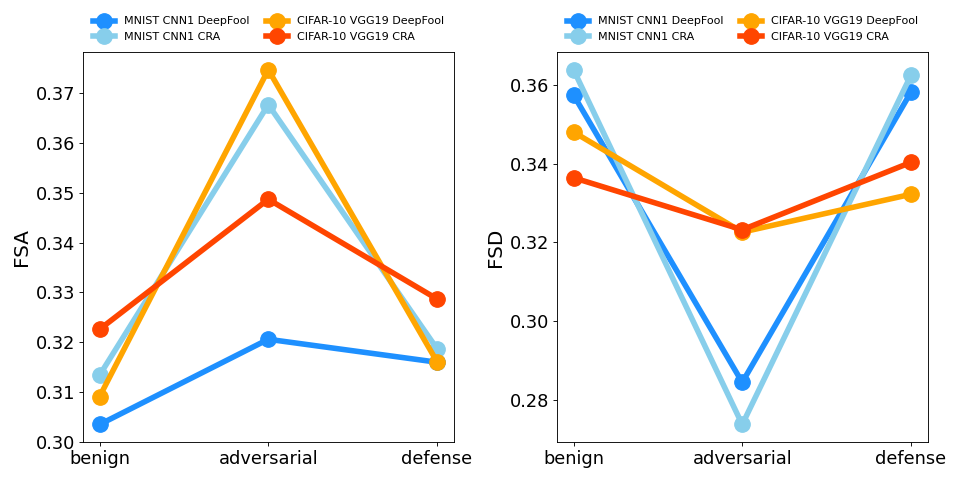}
    \caption{FSA and FSD for benign and adversarial examples before and after defense, where all data in the figure is normalizated.}
    \label{fig:FSAFSD}
\end{figure}

Fig.~\ref{fig:FSAFSD} shows the changes of within-class and among-class distance of four models in MNIST and CIFAR-10 on white-box attack DeepFool and black-box attack CRA. The ordinate represents the value of two distances, and the abscissa represents the output of the benign and adversarial example before and after defense. 

As for within-class distance, eight broken lines show the same trend: all rise first and then decline, with close values in benign and defense columns. The curve of among-class distance showed a “V” shape, which decreases first and then increases. The reason lies in that adversarial attacks increase the distance of within-class, that is to say, scatter the feature distribution of the same class. On the other hand, it also reduces the distance between different classes, making the feature distribution of different classes more concentrated and hard to distinguish. Consequently, misclassification of the model appears. SFE reconstructs the features of adversarial examples to approximate that of benign examples, reduces the distance within class and increases the distance among classes. As the result, it is easier for model to distinguish examples from different classes, and finally guarantees the correct label output by the model after defense.

\subsection{Defense Impact on Benign Examples}
A robust defense should reduce or minimize the misclassification rate caused by various adversarial attacks while maintaining a high classification accuracy. In order to further demonstrate the reliability of SFE, we randomly selected 1000 benign examples from MNIST,CIFAR-10 and ImageNetand input them to the model after defense for re-identification, so as to measure the impact of defense on benign examples. At the same time, we also tested the computation time and obtained results in Table~\ref{tab:benign}.

It can be observed from the table that SFE does not sacrifice the classification accuracy of benign examples while completing efficient defense. It indicates that SFE can accurately find most of the critical pixels that affect the classification results when reconstructing salient features. In this way, no obvious decline in accuracy could be found during the experiment. Besides, SFE consumes less time than that on original setting. This is because SFE extracts features of benign examples for re-identification, decreases the dimension and size of the data to be processed, thus reducing the time cost in the defense process.

\begin{table}[t]
\centering
\caption{The classification accuracy and time complexity of SFE on benign examples before and after defense, where "benign" denotes classification and time results of benign examples while "defense" denotes results of benign examples after SFE defense.}
\label{tab:benign}
\resizebox{0.5\linewidth}{!}{
\begin{tabular}{clcccc}
\hline
\multirow{2}{*}{\textbf{Datasets}}            & \multirow{2}{*}{\textbf{Models}} & \multicolumn{2}{c}{\textbf{acc}}                           & \multicolumn{2}{c}{\textbf{time/s}}                        \\
\cline{3-6}
                                              &                                 & \multicolumn{1}{l}{benign} & \multicolumn{1}{l}{defense} & \multicolumn{1}{l}{benign} & \multicolumn{1}{l}{defense} \\
                                              \cline{1-6}
\multirow{2}{*}{MNIST}                        & CNN1                            & 99.79\%                      & 98.57\%                   & 5.36                         & 5.42                        \\
                                              & CNN2                            & 99.76\%                      & 98.77\%                     & 5.60                         & 5.54                        \\
                                              \cline{2-6}
\multicolumn{1}{l}{\multirow{2}{*}{CIFAR-10}} & AlexNet                         & 99.92\%                      & 98.87\%                    & 6.04                         & 5.95                        \\
\multicolumn{1}{l}{}                          & VGG19                           & 99.88\%                      & 98.53\%                   & 6.15                         & 6.02                        \\\cline{2-6}
\multirow{2}{*}{ImageNet}                     & Inc-v3                          & 99.48\%                      & 98.35\%                    & 7.72                         & 7.60                        \\
                                              & VGG19                           & 99.50\%                       & 91.71\%                     & 7.35                         & 7.28   \\
                                              \hline
\end{tabular}
}
\end{table}

\subsection{Defense Transferability of SFE}
In the real-world scene, defense is carried out without the knowledge of the attack algorithm implemented by the attacker. Therefore, the defense transferability among attacks is particularly important. In this section, we discuss the transferability of SFE. Under this setting, GAN is trained with adversarial examples crafted by a certain attack, but reconstructed SF and TF are used to detect and defend against other attacks. Under this setting, we performed a transferability experiment on SFE to address \textbf{RQ2}.

\begin{figure}[htbp]
  \centering
  \subfigure[DR of MNIST-CNN1]{
  \includegraphics[width=0.48\linewidth]{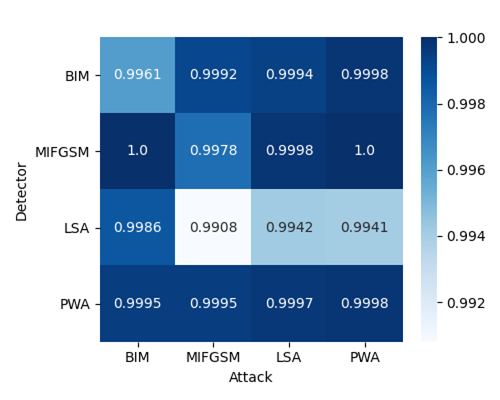} }
  \subfigure[DR of CIFAR-10 AlexNet]{
  \includegraphics[width=0.48\linewidth]{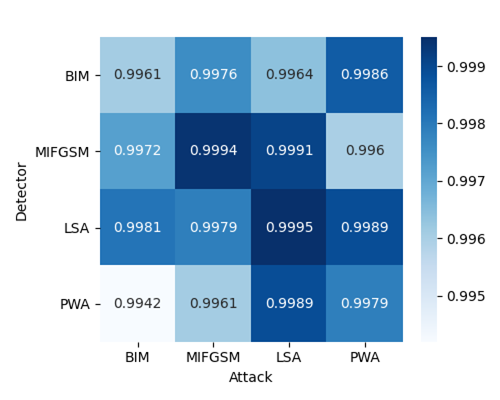} }\\
  \caption{The visualization of heatmap of SFE in detection transferable setting on MNIST and CIFAR-10 dataset, where blue bars denote the results of DR.}
  \label{fig:transfer_det}
\end{figure}

In order to analyze the detection ability of SFE between various attacks, we input adversarial examples into detectors trained by the features of adversarial examples generated by other kinds of attacks for testing. Experiments are carried out in CNN1 of MNIST and AlexNet of CIFAR-10. BIM and MI-FGSM are randomly selected as white-box attacks, with LSA and PWA as black-box attacks. Other experimental settings are the same as that in Section~\ref{detection}. The visualization results of detection are shown in Fig.~\ref{fig:transfer_det}, where the color intensity of the squares is proportional to DR. The horizontal line represents the adversarial examples used for testing, and the vertical line represents those used to train the detector. The result on the diagonal is the non-transferable data in Section~\ref{detection}.

As the figure suggests, SFE shows detection success rate of more than 98\% regardless of whether the salient and trivial features are trained by the detected attack or not. Compared with the adversarial examples of training and transferability testing, they are quite the same in the detection accuracy between them. More specifically, some results of transferability testing even exceed that of the non-transferable experiment. In addition, SFE has good transferability between white-box attacks and black-box attacks, which indicates that the detection performance is fairly independent on adversarial examples used for training.

\begin{figure}[h]
  \centering
  \subfigure[DSR of MNIST-CNN1]{
  \includegraphics[width=0.47\linewidth]{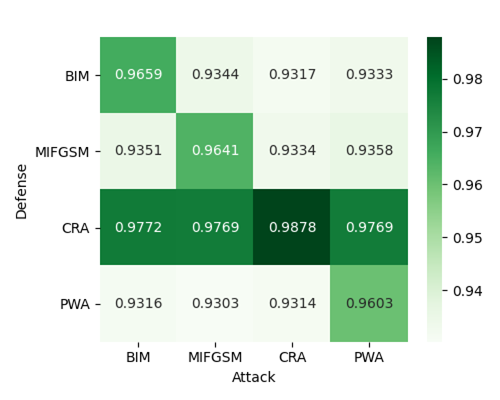} }
  \subfigure[DSR of CIFAR-10 AlexNet]{
  \includegraphics[width=0.47\linewidth]{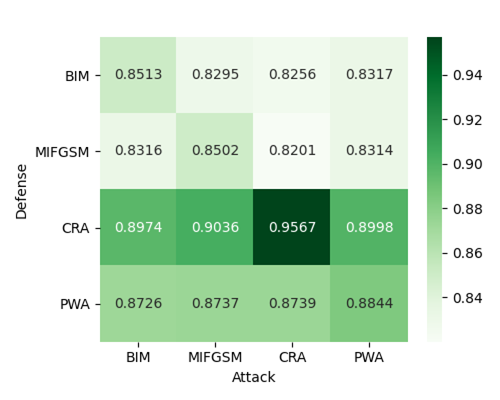} }\\
  \caption{The visualization of heatmap of SFE in detection transferable setting on MNIST and CIFAR-10 dataset, where blue bars denote the results of DR.}
  \label{fig:transfer_def}
\end{figure}

Meanwhile, we also implement similar experiments to investigate the transferability of defense. PWA and CRA are selected as black-box attacks this time, and other experiment settings are consistent with Section~\ref{defense}. Defense results are shown in Fig.~\ref{fig:transfer_def}.

It can be observed from the figure that the results of defense are roughly similar to those of detection. Although the defense effect of the transferability experiment was slightly lower than that of original experiment, no obvious difference between them could be seen from the figure. The same result can be observed as well between black and white-box attacks. No matter what kind of adversarial examples are used for training, SFE can maintain a stable defense effect among a variety of attack methods.

\begin{framed}
For \textbf{RQ2}, the experimental results demonstrate that SFE shows a certain transferability among a variety of attack methods in detection and defense. Moreover, its defensive capability is unrelated to attack algorithms.
\end{framed}

\subsection{Parameter Sensitivity Analysis}
 In this section, to further illustrate the reliability of SFE, we study the influence of hyper-parameter on defense and detection, especially the generator(G) structure of GAN.

\begin{figure}[t]
\centering
    \subfigure[detection results]{
        \includegraphics[width=0.47\linewidth]{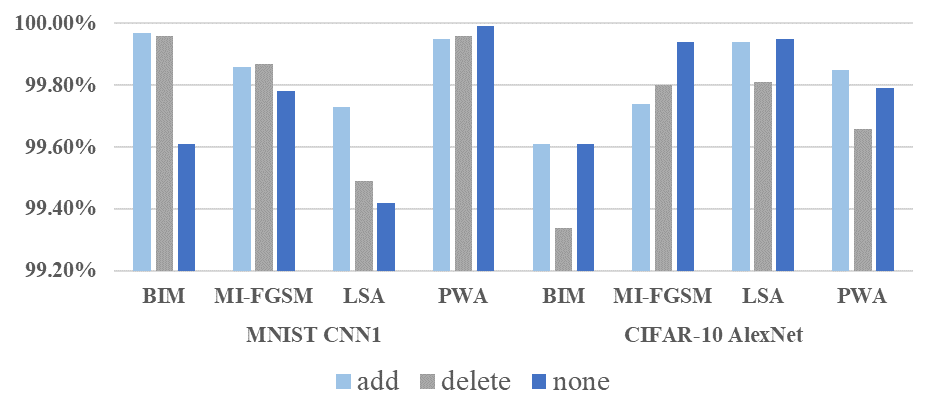}
    }
    \subfigure[defense results]{
        \includegraphics[width=0.47\linewidth]{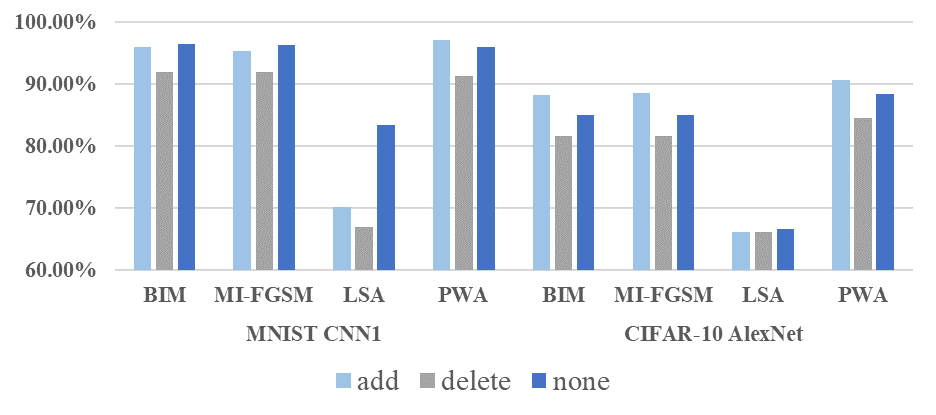}
    }
    \caption{The results of detection and defense on different operations of G structure. This curve is based on the CNN1 on MNIST dataset and AlexNet on CIFAR-10 dataset.
    }
    \label{fig:para}
\end{figure}

We add or delete the layer of G in GAN. The structure of the original G is the same as that described in Section~\ref{method}. We carried out the experiment in CNN1 of MNIST and AlexNet of CIFAR-10. Adversarial examples we used are crafted by BIM, MI-FGSM, LSA, PWA with the same other experimental parameters. The results of detection and defense are shown in Fig.~\ref{fig:para}, where add means adding the layer of G, delete means deleting the layer, and none means unchanged. The structure of the G model after operations are shown in Table~\ref{tab:para_layer}.

\begin{table}[t]
\centering
\caption{The structure of G model after different operations.}
\label{tab:para_layer}
\resizebox{0.45\linewidth}{!}{
\begin{tabular}{lll}
\hline
\multicolumn{1}{c}{\multirow{2}{*}{\textbf{Layers}}} & \multicolumn{2}{c}{\textbf{Operations}} \\\cline{2-3}
\multicolumn{1}{c}{}                                 & add                & delete             \\\cline{1-3}
Dense                                                & 256                & 256                \\
LeakyReLU                                            & alpha=0.2          & alpha=0.2          \\
BatchNormalization                                   & momentum=0.8       & momentum=0.8       \\
Dense                                                & 512                & -                  \\
LeakyReLU                                            & alpha=0.2          & -                  \\
BatchNormalization                                   & momentum=0.8       & -                  \\
Dense                                                & 1024               & 1024               \\
LeakyReLU                                            & alpha=0.2          & alpha=0.2          \\
BatchNormalization                                   & momentum=0.8       & momentum=0.8       \\
Dense                                                & 1024               & -                  \\
LeakyReLU                                            & alpha=0.2          & -                  \\
BatchNormalization                                   & momentum=0.8       & -                  \\
Dense                                                & 1024               & -                  \\
LeakyReLU                                            & alpha=0.2          & -                  \\
BatchNormalization                                   & momentum=0.8       & -    \\\hline
\end{tabular}
}
\end{table}

According to the figure, we can conclude that adding or deleting the structure of G has little effect on the detection effect on MNIST and CIFAR-10 data sets with all DR stable at around more than 99\%. This shows that the difference between salient and trivial feature are still obvious for model to differentiate, which is less susceptible to structure of G. On the other hand, Increasing the structure of G may improve the defense effect to some extent, but deleting the structure of it may slightly reduce the defense effect. This is not hard to understand: By adding the structure of G, the fully connected layer maps the learned distribution features in the high-dimensional space for more times, which improves the learning ability of GAN for salient and trivial features, and vice versa. Meanwhile, we also notice that the simple model like CNN1 is more vulnerable to change of G structure than more complex model like AlexNet in defense, especially in attacks with large perturbation such as LSA.

\subsection{Comparison of Algorithm Complexity}
In the previous section, we have confirmed that SFE has competitive defense capability, good transferability, and stable parameter sensitivity. 

In the design process of the algorithm, effectiveness and efficiency should all be taken into consideration with the fast development of big data and cloud computing technology. Therefore, in this section, the efficiency of the algorithm will be verified. We present a comparison of detection and defense times between SFE and baselines. Combining with the experimental results, we will give an answer to question \textbf{RQ3}. 

\begin{figure}[t]
  \centering
  \subfigure[MNIST-CNN1]{
  \includegraphics[width=0.47\linewidth]{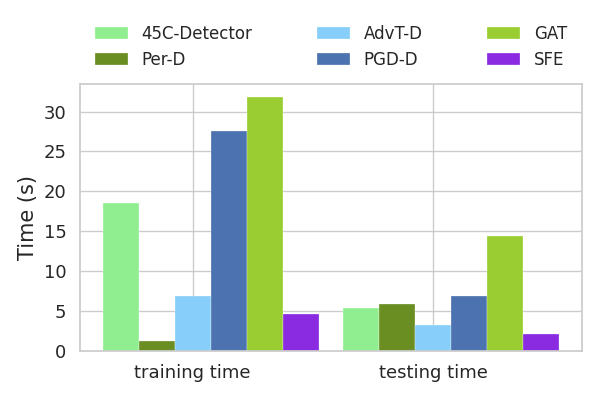}}
  \subfigure[ImageNet Inc-v3]{
  \includegraphics[width=0.47\linewidth]{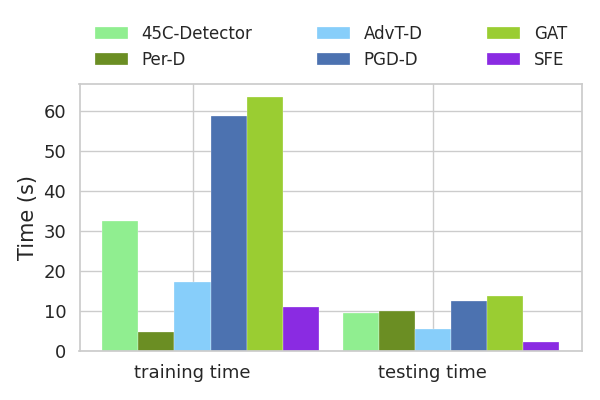}}\\
  \caption{Time comparison among detection algorithms during training and testing for CNN1 classifier on MNIST dataset and Inc-v3 classifier on ImageNet.}
  \label{fig:time_detection}
\end{figure}

In the experiment, We chose 10000 images for training and 1000 for testing. Adversarial examples are crafted by DeepFool. 
Fig.~\ref{fig:time_detection} shows computation time of training and testing during detection on MNIST and ImageNet.

For MNIST, the training time of SFE is almost four times that of Per-D, mainly because GAN needs to learn and imitate the salient and trivial features during training. Compared with the other three detection methods, the training time of SFE is less than 5 seconds, much shorter than others. Besides, the test speed of SFE is the fastest among these detection methods.

Similar pattern could be concluded in the experiments results for ImageNet. But for large data sets, the training complexity of the detector increases, extending the training time. The training time of SFE is still less than 12 seconds, superior to other methods. In terms of testing time, SFE still ranks top, with the shortest time consumption.

\begin{figure}[t]
  \centering
  \subfigure[MNIST-CNN1]{
  \includegraphics[width=0.49\linewidth]{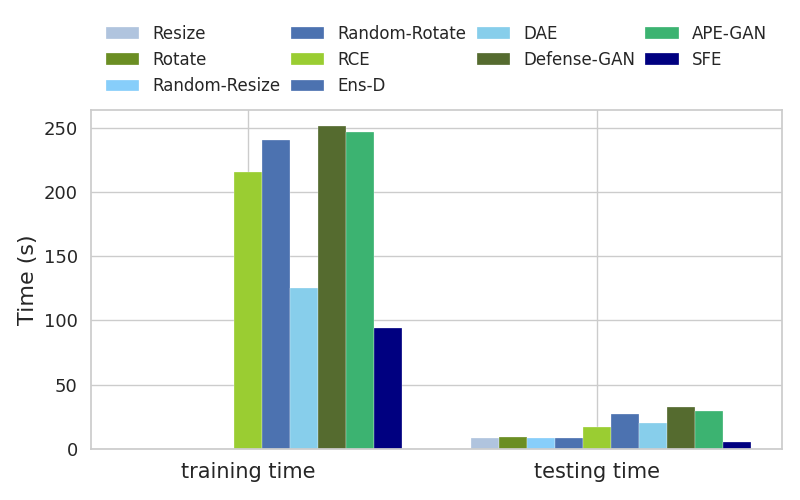}}
  \subfigure[ImageNet Inc-v3]{
  \includegraphics[width=0.49\linewidth]{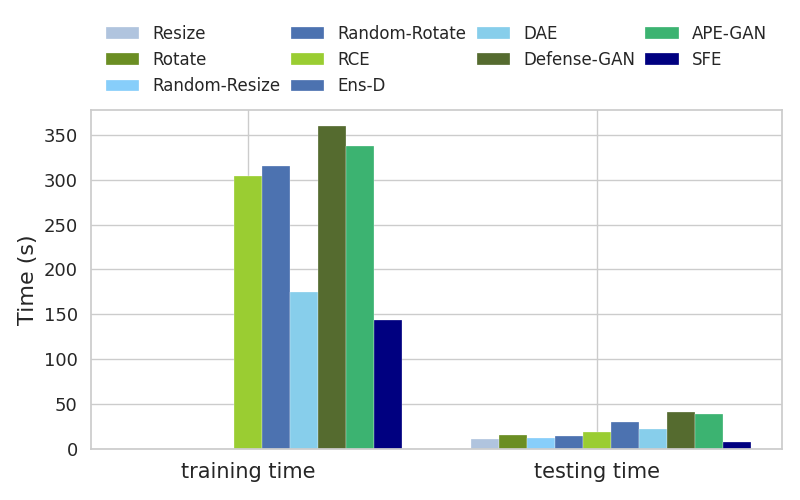}}\\
  \caption{Time comparison among defense algorithms during training and testing for CNN1 classifier on MNIST dataset and Inc-v3 classifier on ImageNet.}
  \label{fig:time_defense}
\end{figure}
We implement the same experiment for defense, the result of which is shown in Fig.~\ref{fig:time_defense}.

As the figure suggests, resize and rotate belong to input transformation defense, which does not involve training operation, so the training time is 0. Ens-D needs to train two ensemble models, so it is time-consuming. RCE, DAE, Defense-GAN and APE-GAN all have to train a new model to filter out perturbations while SFE only needs to reconstruct the features extracted from the model, so the training cost is much lower than these baselines. In general, the test time of all defense methods is quite close, but SFE only needs to re-identify reconstructed features, which proves to be the most efficient algorithm.

\begin{framed}
Here, we can answer \textbf{RQ3}: SFE shows the low cost in terms of time complexity. Concretely, the training time of SFE is less than most of the detection and defense methods that require training. Moreover, the test time is much lower than baselines as well. By comparing and re-identifying features, SFE achieves low complexity and high efficiency, which helps it applicable to various scenarios.
\end{framed}

\subsection{Detection and Defense Results of Adaptive Attack}
In this section, we discuss the detection and defense effectiveness of SFE under adaptive settings, where the attacker knows our defense methods in advance. 

\begin{figure}[t]
  \centering
  \subfigure[MNIST-CNN1]{
  \includegraphics[width=0.15\linewidth]{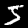}
  \quad
  \includegraphics[width=0.15\linewidth]{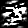}
  }\quad
  \subfigure[CIFAR-10 AlexNet]{
 \includegraphics[width=0.15\linewidth]{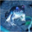}
  \quad
  \includegraphics[width=0.15\linewidth]{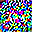}}\\
  \caption{Adaptive adversarial examples of MNIST and CIFAR-10, where the left is benign example while the right is its adversarial example.}
  \label{fig:adaptive_example}
\end{figure}

To prevent SFE from extracting and separating SF and TF from the input, we increased the perturbation size added to the benign examples to 0.08, almost 100 times that of the previous experiment. The specific adversarial examples on MNIST and CIFAR-10 are shown in Fig.~\ref{fig:adaptive_example}, where a human can hardly tell the contents of the image. The results of detection and defense, as shown in Table~\ref{tab:adaptive}, are obtained with the same experimental parameters as Section~\ref{detection} and Section~\ref{defense}.

From the data in table~\ref{tab:adaptive}, SFE still has certain detection and defense capabilities for adversarial examples with such exaggerated perturbations: more than half of adversarial examples can be detected and finally re-identified. It is indicated that SFE shows robustness under adaptive attack settings. Generally speaking, for adaptive attacks, detection results of SFE are better than defense. Reasons lie in that detection only calculates the difference between SF and TF, and does not require the label-related information in extracted features. On the contrary, re-identification defense directly use that information of important features for classification, which is more sensitive to the large perturbation added by adaptive attacks in the image.

\begin{table}[htbp]
\centering
\caption{The detection and defense performance of SFE under adaptive attack settings.}
\label{tab:adaptive}
\resizebox{0.5\linewidth}{!}{
\begin{tabular}{l|cc|ll|ll}
\hline
\textbf{Datasets} & \multicolumn{2}{c}{MNIST} & \multicolumn{2}{c}{CIFAR-10}                             & \multicolumn{2}{c}{ImageNet}                            \\
\hline
\textbf{Model}    & CNN1        & CNN2        & \multicolumn{1}{c}{AlexNet} & \multicolumn{1}{c|}{VGG19}  & \multicolumn{1}{c}{Inc-v3} & \multicolumn{1}{c}{VGG19}  \\\hline
\textbf{DR}       & 70.37\%      & 71.52\%      & \multicolumn{1}{c}{70.40\%}  & \multicolumn{1}{c|}{70.20\%} & \multicolumn{1}{c}{68.60\%} & \multicolumn{1}{c}{69.20\%} \\
\textbf{DSR}      & 56.10\%      & 54.96\%      & \multicolumn{1}{c}{55.92\%}  & \multicolumn{1}{c|}{52.04\%} & \multicolumn{1}{c}{55.06\%} & \multicolumn{1}{c}{58.64\%}
 \\\hline
\end{tabular}
}
\end{table}

\subsection{Visualization based Interpretation}
In this section, we will analyze and verify the effectiveness of defense and detection from high-dimensional feature space via t-SNE~\cite{van2008visualizing} and image pixel features via Grad-CAM~\cite{selvaraju2017grad}. Interpretable understanding of the whole process is detailed and \textbf{RQ4} is also answered later.
\subsubsection{Visualization of t-SNE}
t-SNE is adopted to visualize the process of attack, defense and detection from the perspective of cluster distribution in high-dimensional feature space.

\textcircled{1} Visualization of Attacks
\begin{figure}[htbp]
  \centering
  \subfigure[benign]{
  \includegraphics[width=0.2\linewidth]{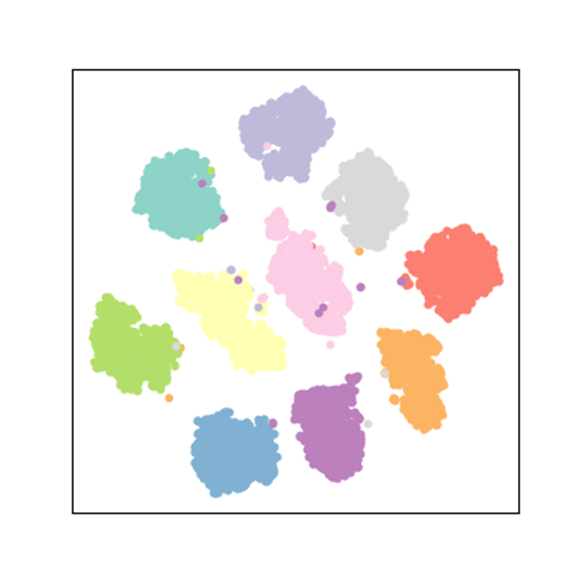} }
  \subfigure[MI-FGSM]{
  \includegraphics[width=0.2\linewidth]{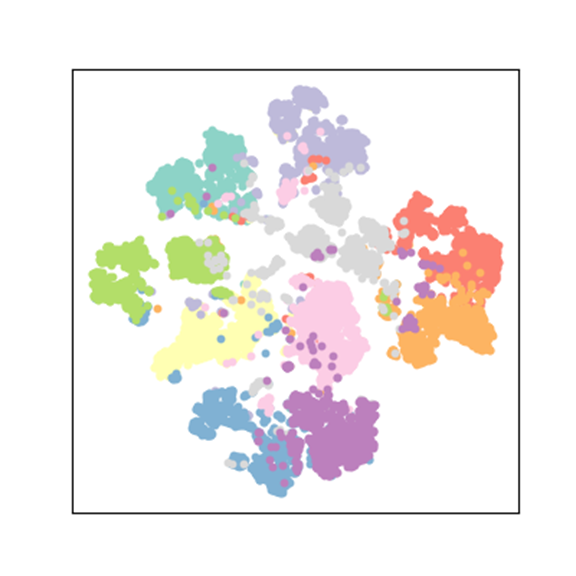} }
  \subfigure[JSMA]{
  \includegraphics[width=0.2\linewidth]{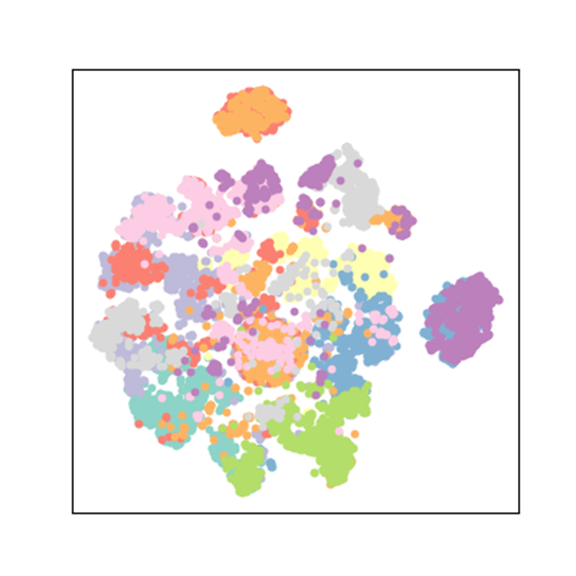} }
  \subfigure[PWA]{
  \includegraphics[width=0.2\linewidth]{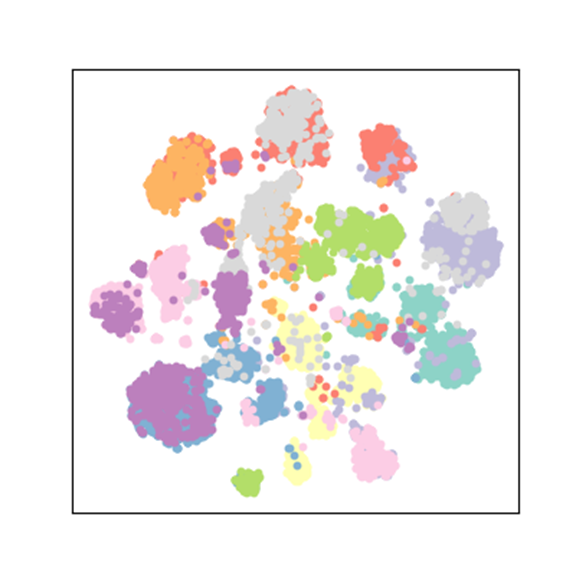} }\\
  \caption{The t-NSE visualization of different attacks on CNN1 of MNIST dataset, where the last three images are  high-dimensional feature distributions of the adversarial examples crafted by MI-FGSM, JSMA, PWA respectively.}
  \label{fig:tsne_attack}
\end{figure}

Fig.~\ref{fig:tsne_attack} shows the t-NSE visualization results of benign examples before and after attack on MNIST-CNN1 model. From left to right, are corresponding t-NSE results of benign examples, and adversarial examples generated by MI-FGSM, JSMA and PWA. Different colour blocks represent different class clusters.

The well-trained model can accurately classify images from different class, so all kinds of clusters in t-SNE visualization of benign examples are separated from each other. Comparing the visualization result of the benign and adversarial example, we can find that the attacks break up the distribution of classes while cutting down the distance between different classes. Whether it is the gradient-based white-box attack MI-FGSM or decision- based black-box attack PWA, they all fool the model by increasing the distance between classes.

\textcircled{2} SF and TF of benign example
\begin{figure}[htbp]
  \centering
  \subfigure[benign]{
  \includegraphics[width=0.2\linewidth]{tsne/CNN1/ori.png} }
  \subfigure[SF of benign]{
  \includegraphics[width=0.2\linewidth]{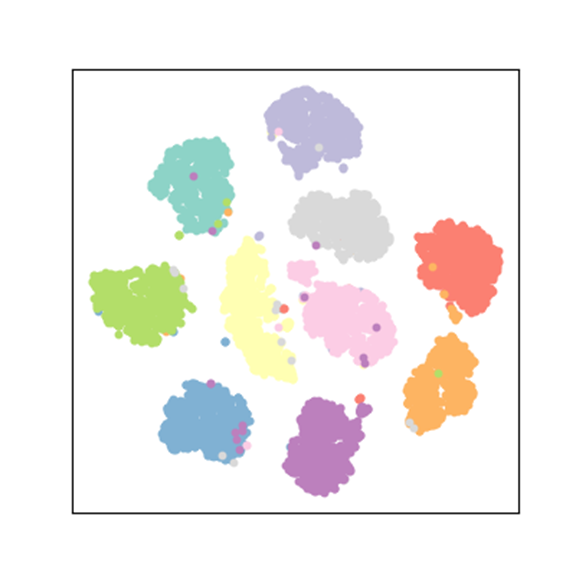} }
  \subfigure[TF of benign]{
  \includegraphics[width=0.2\linewidth]{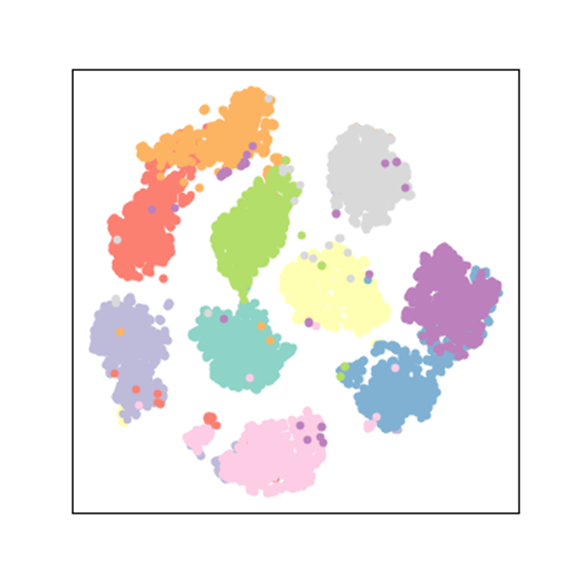} }
  \\
  \caption{The t-NSE visualization of SF and TF of benign examples on MNIST CNN1.}
  \label{fig:tsne_SFTF}
\end{figure}

Fig.~\ref{fig:tsne_SFTF} shows the t-SNE visualization of benign examples and their corresponding SF and TF on MNIST-CNN1. From left to right are benign examples classified by the original model, SF and TF reconstructed by GAN.

It can be seen from the figure that the distribution of SF and TF of benign examples after reconstruction of GAN are similar, both scattered, which is consistent with our definition in Section~\ref{pre}. The distribution of the reconstructed SF resembles that of the original model, which indicates that the defense poses little negative impact on classification accuracy of benign examples.

\textcircled{3} The effectiveness of detection
\begin{figure}[htbp]
  \centering
  \subfigure[SF of benign]{
  \includegraphics[width=0.2\linewidth]{tsne/CNN1/ori_sf.png} }
  \subfigure[TF of benign]{
  \includegraphics[width=0.2\linewidth]{tsne/CNN1/ori_tf.png} }
  \subfigure[SF of adversarial]{
  \includegraphics[width=0.2\linewidth]{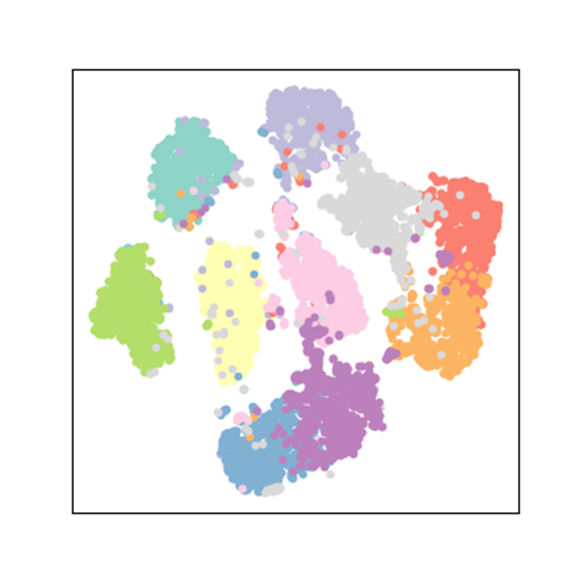} }
  \subfigure[TF of adversarial]{
  \includegraphics[width=0.2\linewidth]{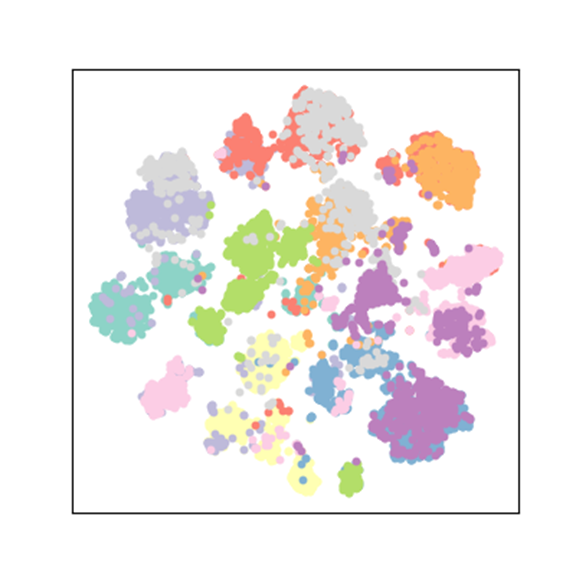} }
  \\
  \caption{Reconstructed SF and TF of benign and adversarial examples on MNIST-CNN1. Adversarial examples are generated by PWA.}
  \label{fig:tsne_detect}
\end{figure}

Fig.~\ref{fig:tsne_detect} shows SF and TF of benign and adversarial examples on the MNIST-CNN1 model. 
As can be seen from the figure, the distribution of SF and TF of benign examples looks similar, while that of adversarial examples are quite different, which conforms to our previous definition as well. SFE detectors learn the differences between SF and TF, thus successfully detecting adversarial examples.

\textcircled{4} The effectiveness of defense
\begin{figure}[htbp]
  \centering
  \subfigure[benign]{
  \includegraphics[width=0.2\linewidth]{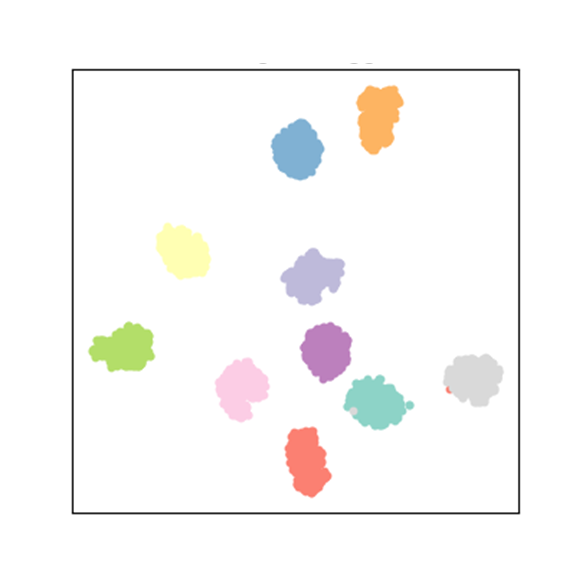} }
  \subfigure[MI-FGSM attack]{
  \includegraphics[width=0.2\linewidth]{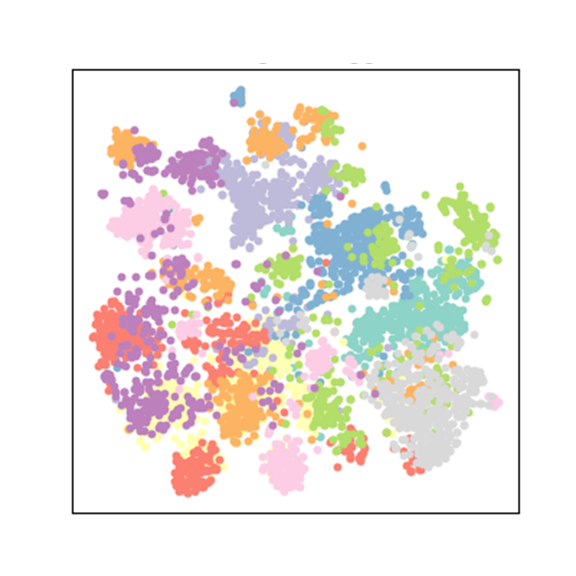} }
  \subfigure[SF of adversarial]{
  \includegraphics[width=0.2\linewidth]{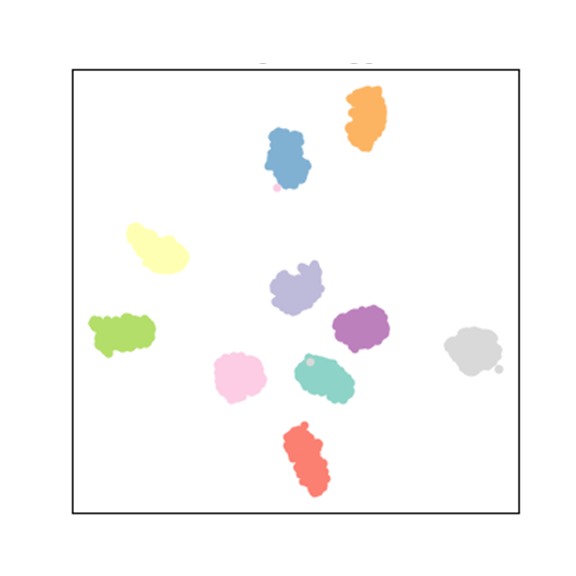} }
  \subfigure[SF of benign]{
  \includegraphics[width=0.2\linewidth]{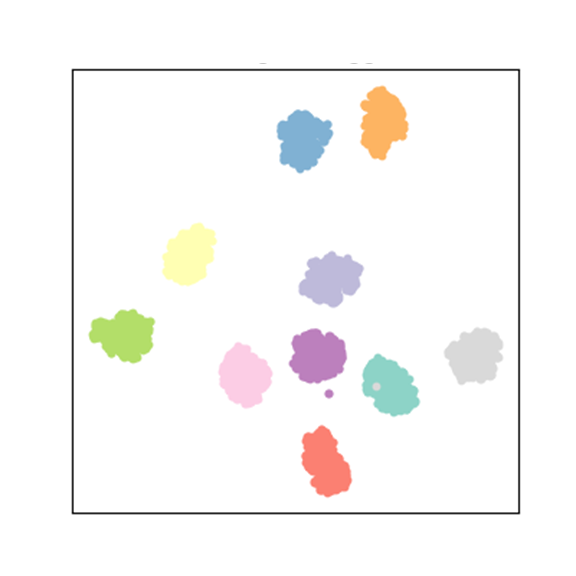} }
  \\
  \caption{t-NSE visualization of ImageNet VGG 19 during the defense process. Adversarial examples are crafted by MI-FGSM.}
  \label{fig:tsne_defense}
\end{figure}

Fig.~\ref{fig:tsne_defense} shows the visualization results of VGG19 model on ImageNet before and after attack and after defense. From left to right are benign and adversarial examples crafted by MI-FGSM attack, SF of adversarial examples after GAN reconstruction and SF of reconstructed benign image.

After reconstructing by GAN, the distribution of t-SNE of benign and adversarial examples are similar to that of original model. The visualization result demonstrates the GAN's competitive performance of learning in terms of feature distribution: it will study the features of the adversarial examples to resemble SF of benign examples, so that the reconstructed SF of adversarial examples will be classified correctly when input to the model for re-identification defense. Besides, the SF of adversarial examples after reconstruction of GAN is similar to that of the benign example, consistent with the definition proposed above, which also verifies the validity of re-identification of SF during defense.

\textcircled{5} The effectiveness of defense

From the results of t-SNE visualization above, it can be concluded that images of a certain class are misclassified to another class by attacks. During that process, the cluster distribution is disrupted and the distance between classes is reduced. SFE remaps the distribution of different features, widens the distance between classes, and scatters the distribution, approximating the original model. Consequently, defense and detection of SFE are based on the characteristics of class distribution and independent of attack methods, which contributes to transferability among different attack algorithms.

\subsubsection{Visualization of Grad-CAM}

\begin{figure}[htbp]
  \centering
  \subfigure[benign example]{
  \includegraphics[width=0.18\linewidth]{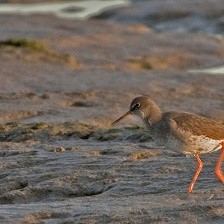} }
  \subfigure[heatmap of benign]{
  \includegraphics[width=0.18\linewidth]{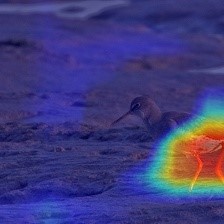} }
  \subfigure[SF of benign]{
  \includegraphics[width=0.18\linewidth]{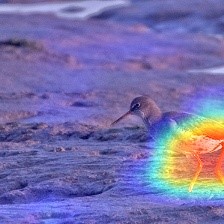} }
  \subfigure[TF of benign]{
  \includegraphics[width=0.18\linewidth]{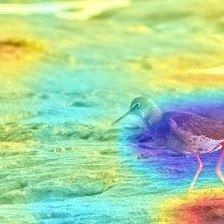} }
  \\
  \subfigure[adversarial example]{
  \includegraphics[width=0.18\linewidth]{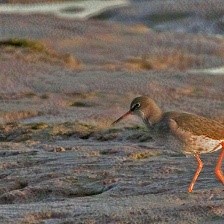} }
  \subfigure[heatmap of adv]{
  \includegraphics[width=0.18\linewidth]{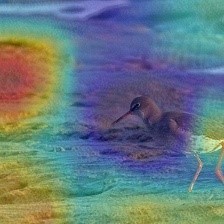} }
  \subfigure[SF of adv]{
  \includegraphics[width=0.18\linewidth]{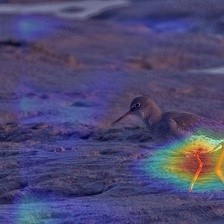} }
  \subfigure[TF of adv]{
  \includegraphics[width=0.18\linewidth]{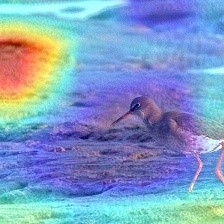} }\\
  \subfigure[benign example]{
  \includegraphics[width=0.18\linewidth]{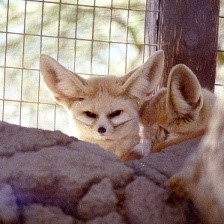} }
  \subfigure[heatmap of benign]{
  \includegraphics[width=0.18\linewidth]{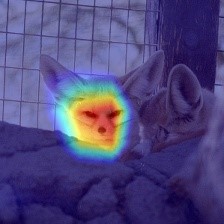} }
  \subfigure[SF of benign]{
  \includegraphics[width=0.18\linewidth]{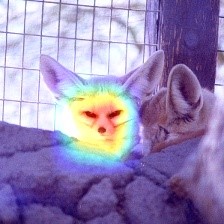} }
  \subfigure[TF of benign]{
  \includegraphics[width=0.18\linewidth]{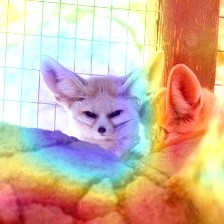} }
  \\
  \subfigure[adversarial example]{
  \includegraphics[width=0.18\linewidth]{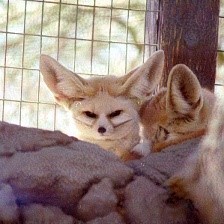} }
  \subfigure[heatmap of adv]{
  \includegraphics[width=0.18\linewidth]{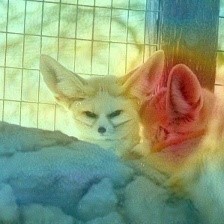} }
  \subfigure[SF of adv]{
  \includegraphics[width=0.18\linewidth]{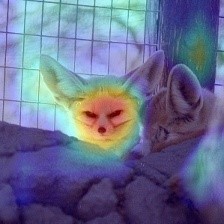} }
  \subfigure[TF of adv]{
  \includegraphics[width=0.18\linewidth]{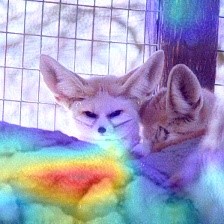} }\\
  \caption{Visualization of SF and TF on ImageNet VGG19 model via Grad-CAM. From left to right are the benign example and its heatmap, visualization of SF and TF reconstructed by SFE. The second row represents heatmaps of adversarial examples  and their corresponding SF and TF results after reconstruction. From red to blue, the weight that the model allocates decreases. Adversarial examples are generated by DeepFool and BIM, respectively.}
  \label{fig:heatmaps}
\end{figure}

Grad-CAM provides heatmap visualization from the perspective of pixel-level features. Here, we provide more heatmaps generated by Grad-CAM on ImageNet for detailed indication, as shown in Figure \ref{fig:heatmaps}. Areas more closely related to classification label and drawn more attention by the model are painted in red. From red to blue, the weight that the model allocates decreases. More visualization results will be shown in \ref{appendix_visualization}.

It can be easily observed from the figure that red areas in  SF of benign are quite similar to that in heatmap of benign example. This demonstrates that SF is relevant to class label, consistent with our definition and human semantics. Therefore, defense effect is guaranteed by the high similarity of highlight areas between reconstructed SF of adversarial examples and that in benign ones. On the contrary, TF are class-independent features. When attention is paid to those areas irrelevant to the label, misclassification happens, as shown in heatmap of adversarial. 

\begin{framed}
Based on the visualization results of t-SNE and Grad-CAM, we can answer \textbf{RQ4} with the following conclusions: 1) In the visualization of the high-dimensional features of the model, SFE successfully achieved the defense against the adversarial examples by reconstructing the feature distribution and increasing the distance between classes; 2) As for pixel-level feature in an image, SF is quite class-related while TF may cause misclassification. By reconstructing SF of adversarial exmaples, the defense effect is reached when it resembles a benign one. 
\end{framed}

\section{Conclusions\label{conclusion}}
In this paper, we propose the concepts of salient features and trivial features. We use GAN framework to extract these two features, and put forward the SFE method, which can detect and re-identify adversarial examples at the same time. Comprehensive experiments have shown that, compared with baselines, SFE achieves fairly good detection and defense effect for a variety of attack algorithms on different datasets and models, and has fairly good defense transferability among different attacks.

To improve the robustness of DNNs against attacks, as a future work, we plan further to improve the defense accuracy of SFE on complex datasets by optimizing the network structure and training strategy. Meanwhile, we would cut down the complexity of model structure and computation time to meet large-scale applications and computing needs.

\section*{Acknowledgment}
This research was supported by 
the National Natural Science Foundation of China under Grant No. 62072406, 
the Natural Science Foundation of Zhejiang Provincial under Grant No. LY19F020025, 
the Major Special Funding for ``Science and Technology Innovation 2025'' in Ningbo under Grant No. 2018B10063, 
the National Key Research and Development Program of China under Grant No. 2018AAA0100801,
the Key Laboratory of the Public Security Ministry Open Project in 2020 under Grant No. 2020DSJSYS001.

% \bibliography{myref}

% \bibliography{myref}

\newpage

\appendix

\section{More visualization of SF and TF \label{appendix_visualization}}

\subsection{Visualization of CIFAR-10}
Here we show visualizations of CIFAR-10 on VGG19 model. From left to right are the benign example and its heatmap of SF and TF in adversarial exmaples. From red to blue, the weight that the model allocates decreases. Adversarial examples are generated by all attacks used in Section \ref{setup}.
\begin{figure}[htbp]
  \centering
  \subfigure[benign example]{
  \includegraphics[width=0.15\linewidth]{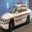} }
  \subfigure[SF of adv]{
  \includegraphics[width=0.15\linewidth]{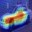} }
  \subfigure[TF of adv]{
  \includegraphics[width=0.15\linewidth]{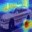} }
  \subfigure[benign example]{
  \includegraphics[width=0.15\linewidth]{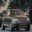} }
  \subfigure[SF of adv]{
  \includegraphics[width=0.15\linewidth]{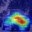} }
  \subfigure[TF of adv]{
  \includegraphics[width=0.15\linewidth]{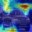} }
  \\
   \subfigure[benign example]{
  \includegraphics[width=0.15\linewidth]{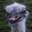} }
  \subfigure[SF of adv]{
  \includegraphics[width=0.15\linewidth]{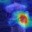} }
  \subfigure[TF of adv]{
  \includegraphics[width=0.15\linewidth]{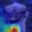} }
  \subfigure[benign example]{
  \includegraphics[width=0.15\linewidth]{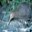} }
  \subfigure[SF of adv]{
  \includegraphics[width=0.15\linewidth]{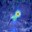} }
  \subfigure[TF of adv]{
  \includegraphics[width=0.15\linewidth]{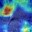} }\\
   \subfigure[benign example]{
  \includegraphics[width=0.15\linewidth]{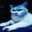} }
  \subfigure[SF of adv]{
  \includegraphics[width=0.15\linewidth]{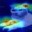} }
  \subfigure[TF of adv]{
  \includegraphics[width=0.15\linewidth]{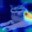} }
  \subfigure[benign example]{
  \includegraphics[width=0.15\linewidth]{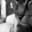} }
  \subfigure[SF of adv]{
  \includegraphics[width=0.15\linewidth]{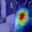} }
  \subfigure[TF of adv]{
  \includegraphics[width=0.15\linewidth]{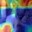} }\\
  \subfigure[benign example]{
  \includegraphics[width=0.15\linewidth]{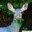} }
  \subfigure[SF of adv]{
  \includegraphics[width=0.15\linewidth]{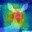} }
  \subfigure[TF of adv]{
  \includegraphics[width=0.15\linewidth]{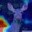} }
  \subfigure[benign example]{
  \includegraphics[width=0.15\linewidth]{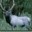} }
  \subfigure[SF of adv]{
  \includegraphics[width=0.15\linewidth]{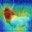} }
  \subfigure[TF of adv]{
  \includegraphics[width=0.15\linewidth]{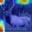} }\\
   \subfigure[benign example]{
  \includegraphics[width=0.15\linewidth]{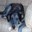} }
  \subfigure[SF of adv]{
  \includegraphics[width=0.15\linewidth]{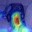} }
  \subfigure[TF of adv]{
  \includegraphics[width=0.15\linewidth]{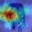} }
  \subfigure[benign example]{
  \includegraphics[width=0.15\linewidth]{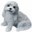} }
  \subfigure[SF of adv]{
  \includegraphics[width=0.15\linewidth]{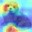} }
  \subfigure[TF of adv]{
  \includegraphics[width=0.15\linewidth]{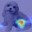} }\\
  \caption{Visualization of SF and TF on CIFAR-10 VGG19 model via Grad-CAM. }
\end{figure}

\subsection{Visualization of ImageNet}
The supplementary visualization of SF and TF on ImageNet VGG19 model via Grad-CAM are shown as follows. From left to right are the benign example and its heatmap of SF and TF in adversarial exmaples. From red to blue, the weight that the model allocates decreases. Adversarial examples are generated by all attacks used in Section \ref{setup}.
\begin{figure}[htbp]
  \centering
  \subfigure[benign example]{
  \includegraphics[width=0.15\linewidth]{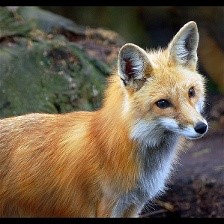} }
  \subfigure[SF of adv]{
  \includegraphics[width=0.15\linewidth]{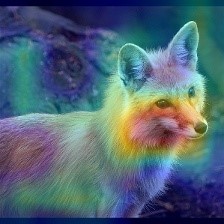} }
  \subfigure[TF of adv]{
  \includegraphics[width=0.15\linewidth]{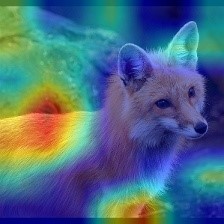} }
  \subfigure[benign example]{
  \includegraphics[width=0.15\linewidth]{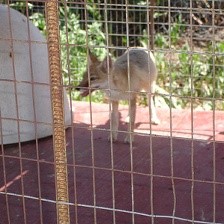} }
  \subfigure[SF of adv]{
  \includegraphics[width=0.15\linewidth]{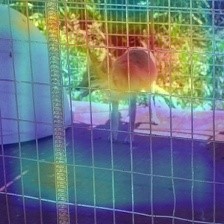} }
  \subfigure[TF of adv]{
  \includegraphics[width=0.15\linewidth]{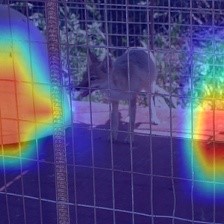} }
  \\
  \subfigure[benign example]{
  \includegraphics[width=0.15\linewidth]{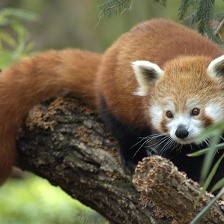} }
  \subfigure[SF of adv]{
  \includegraphics[width=0.15\linewidth]{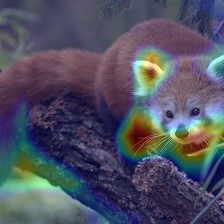} }
  \subfigure[TF of adv]{
  \includegraphics[width=0.15\linewidth]{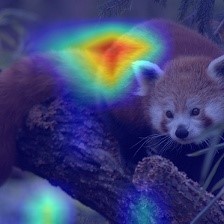} }
  \subfigure[benign example]{
  \includegraphics[width=0.15\linewidth]{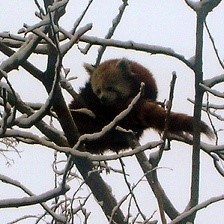} }
  \subfigure[SF of adv]{
  \includegraphics[width=0.15\linewidth]{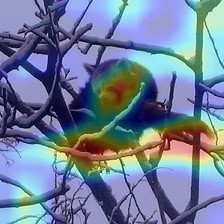} }
  \subfigure[TF of adv]{
  \includegraphics[width=0.15\linewidth]{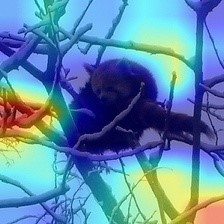} }\\
   \subfigure[benign example]{
  \includegraphics[width=0.15\linewidth]{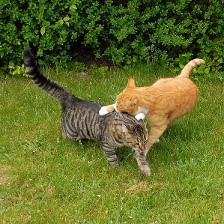} }
  \subfigure[SF of adv]{
  \includegraphics[width=0.15\linewidth]{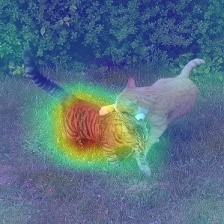} }
  \subfigure[TF of adv]{
  \includegraphics[width=0.15\linewidth]{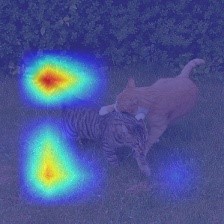} }
  \subfigure[benign example]{
  \includegraphics[width=0.15\linewidth]{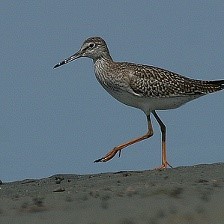} }
  \subfigure[SF of adv]{
  \includegraphics[width=0.15\linewidth]{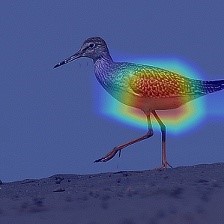} }
  \subfigure[TF of adv]{
  \includegraphics[width=0.15\linewidth]{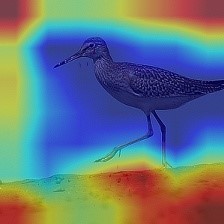} }\\
  \subfigure[benign example]{
  \includegraphics[width=0.15\linewidth]{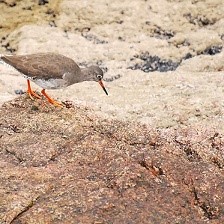} }
  \subfigure[SF of adv]{
  \includegraphics[width=0.15\linewidth]{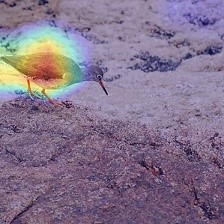} }
  \subfigure[TF of adv]{
  \includegraphics[width=0.15\linewidth]{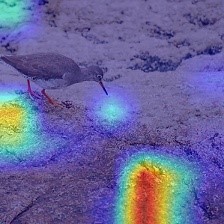} }
  \subfigure[benign example]{
  \includegraphics[width=0.15\linewidth]{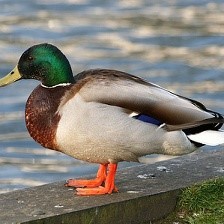} }
  \subfigure[SF of adv]{
  \includegraphics[width=0.15\linewidth]{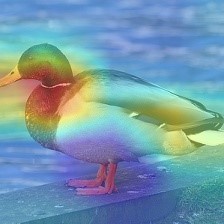} }
  \subfigure[TF of adv]{
  \includegraphics[width=0.15\linewidth]{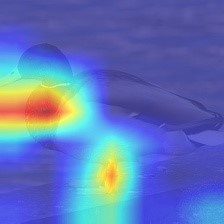} }\\
   \subfigure[benign example]{
  \includegraphics[width=0.15\linewidth]{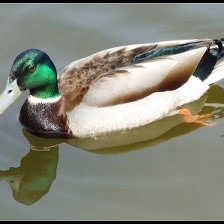} }
  \subfigure[SF of adv]{
  \includegraphics[width=0.15\linewidth]{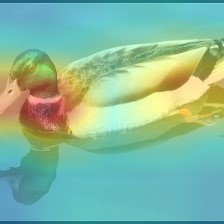} }
  \subfigure[TF of adv]{
  \includegraphics[width=0.15\linewidth]{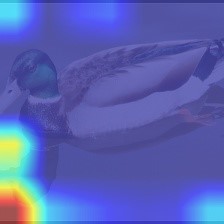} }
  \subfigure[benign example]{
  \includegraphics[width=0.15\linewidth]{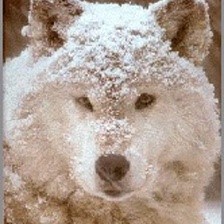} }
  \subfigure[SF of adv]{
  \includegraphics[width=0.15\linewidth]{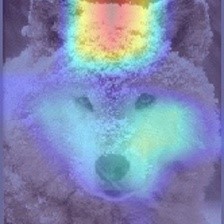} }
  \subfigure[TF of adv]{
  \includegraphics[width=0.15\linewidth]{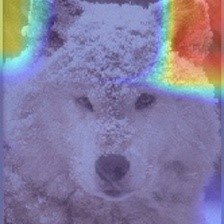} }\\
  \caption{Visualization of SF and TF on ImageNet VGG19 model via Grad-CAM. }
\end{figure}

\end{document}